\documentclass{article}

\usepackage[position, final]{neurips_2026}

\usepackage[utf8]{inputenc}
\usepackage[T1]{fontenc}
\usepackage[hypertexnames=false]{hyperref}
\usepackage{url}
\usepackage{booktabs}
\usepackage{amsfonts}
\usepackage{amsmath}
\usepackage{amssymb}
\usepackage{nicefrac}
\usepackage{microtype}
\usepackage{xcolor}
\usepackage{graphicx}
\usepackage{subcaption}
\usepackage{xspace}
\usepackage{tabularx}
\usepackage{multirow}
\usepackage{enumitem}
\usepackage{float}
\usepackage{wrapfig}
\usepackage[most]{tcolorbox}
\usepackage{algorithm}
\usepackage{algpseudocode}
\usepackage{tikz}
\usepackage{pgfplots}

\title{Robots Need More Than VLAs \& World Models}

\author{%
  Elis Karcini\\
  Motoniq.ai\\
  \And 
  Faisal Mehrban \\
  Motoniq.ai\\
  \And
  Quang Nguyen \\
  Motoniq.ai
 \And 
  Mac Schwager\\
  Stanford University\\
  Motoniq.ai
  \And 
  Arash Ajoundani\\
  Istituto Italiano di Tecnologia
  \And 
  César Cadena \\
  ETH Zurich \\
  \And 
  Jan Peters \\
  Technical University of Darmstadt \\
  \And 
  Marco Hutter \\
  ETH Zurich \\
  \And 
  Haitham Bou-Ammar \\
  UCL Centre for AI 
}
\begin{document}

\maketitle

\begin{abstract}
Generalist robot intelligence is often framed as a policy-scaling problem: collect more robot demonstrations, train larger Vision-Language-Action (VLA) models, and expect broader generalisation. In this position paper, we argue that this framing is incomplete. The central bottleneck is not only policy learning, but the absence of mechanisms that convert the world’s abundant unstructured behavioural data into grounded robot supervision. Human motion, internet video, simulation rollouts, and interactive demonstrations contain rich information about tasks, goals, contacts, failures, and physical constraints, yet most of this information is not directly usable by robot policies because it lacks embodiment-specific action labels, task semantics, and reward structure. We identify four missing components for the next generation of robotics: data interfaces for autolabelling unstructured behaviour, embodiment interfaces for retargeting human motion to robot actions, world-model interfaces for physics-grounded 3D reasoning, and reward interfaces for inferring task progress and success from video and language. We survey recent progress in robot foundation models, cross-embodiment datasets, learning from video, world models, and reward modelling, and propose a research agenda for building robotics systems that can learn not only from robot demonstrations, but from the broader physical world. 
\end{abstract}

\section{Introduction}
\label{sec:introduction}
Robotics is entering its foundation-model moment, but it does not yet have its internet. Large-scale vision-language-action models, cross-embodiment datasets, learned simulators, and interactive world models have made it increasingly plausible that robots will eventually acquire broad, reusable physical skills rather than being programmed task by task. Yet the path to generalist robotics remains far less clear than the path that enabled progress in language and vision. Text and images are abundant, naturally digitised, and densely associated with human-generated supervision. Physical interaction is different: the world contains vast amounts of behavioural data, but most of it is not directly usable by robots. Human demonstrations, internet videos, factory workflows, household activities, and simulation rollouts often reveal what task is being attempted, how objects move, when contact occurs, and whether an outcome succeeds or fails. However, they rarely provide the embodiment-specific action labels, force signals, task semantics, or reward structure required to train robot policies. The central bottleneck for future robotics is therefore not only how to scale policies, but how to convert unstructured physical experience into grounded robot supervision.

Recent progress has begun to address this bottleneck by scaling robot datasets, pooling experience across embodiments, and training generalist vision-language-action policies. These efforts are important: they show that robot behaviour can improve when models are exposed to more tasks, more environments, and more bodies. However, they also reveal a deeper limitation. Most current pipelines still depend on explicitly collected robot demonstrations, manually specified tasks, curated datasets, or embodiment-specific action spaces. This makes progress expensive and difficult to scale. A robot dataset is not like a text corpus: every trajectory must be physically executable, every action is tied to a particular body, and every failure may damage hardware, objects, or the environment. As a result, the amount of usable robot supervision remains tiny compared with the amount of physical behaviour already present in the world. The key question is therefore not only how to collect more robot data, but how to make broader sources of physical experience usable for robot learning.

In this position paper, we survey the rapidly growing body of work that attempts to widen the sources of supervision for robot learning, including vision-language-action models trained on heterogeneous robot trajectories, methods that extract behavioural priors from human and internet-scale video, action-conditioned world models for imagined robot experience, and simulation-based pipelines for generating counterfactual interaction data. Rather than organising this progress only by data source or algorithmic family, we argue that a more useful organising principle is the robotics learning pipeline itself. Today’s pipeline is largely robot-data-centric: collect robot demonstrations, attach task or language labels, train a policy, evaluate on hardware, and repeat. We argue that the future pipeline must instead be grounding-centric: start from broad physical experience, e.g., human motion, internet video, robot interaction, simulation, tactile sensing, and language and pass it through grounding mechanisms that produce robot-usable actions, contacts, object states, task phases, goals, and rewards. The central question for the field is therefore not simply which model architecture to train, but which mechanisms are needed to make the world’s physical experience learnable by robots. Importantly, our argument is not that VLA models are unimportant. Rather, VLAs should be understood as one layer in a larger physical-intelligence stack: a policy interface that depends on upstream grounding of data, embodiment, dynamics, rewards, and deployment feedback.

We therefore organise the survey not merely by model family, dataset, or algorithmic trend, but by the supervision bottleneck each line of work exposes. Robot-native datasets show how far policy learning can scale when actions and task labels are already available. Video-based methods show that the world contains abundant behavioural evidence, but weak grounding. Simulation and world models show how experience can be generated, but only when physical consequences are preserved. This organisation leads to our central claim: the missing layer in robotics is not another policy architecture alone, but a set of components that transform physical experience into robot-usable supervision. 

We argue that the central question for the field is not simply which model architecture to train, but which mechanisms are needed to make the world’s physical experience learnable by robots. We argue that this requires four missing pillars: a physical data engine for embodied autolabelling, task-preserving retargeting across embodiments, physics-grounded world models for consequence prediction, and task-conditioned reward grounding through deployment loops. These pillars provide the organising principle for the remainder of the paper: we first survey how current robot-learning systems expose the grounding bottleneck, and then outline the components needed to move from physical experience to physical intelligence.

\section{Robot-Native Supervision: Progress and Scaling Limits}
Much of contemporary robot learning remains organised around robot-native supervision. A robot is placed in an environment, demonstrations or interaction trajectories are collected, and the resulting observations are paired with embodiment-specific actions, task labels, language instructions, or success signals. This paradigm has enabled substantial progress in imitation learning, reinforcement learning, and vision-language-action modelling, especially as datasets have expanded across tasks, environments, and robot embodiments. At the same time, the field is already searching for ways to move beyond this regime: simulation is used to generate scalable experience, real-to-sim-to-real methods aim to amplify scarce real data, human and internet videos are used to learn behavioural priors, and world models are trained to support prediction, planning, and counterfactual reasoning. Our point is therefore not that robotics is limited to one pipeline, but that much of today’s usable supervision still becomes useful only after it has been grounded into robot-native quantities. The central scaling limit is this grounding step: how do we turn broader physical experience into actions, contacts, object states, task phases, goals, and rewards that a robot can learn from?
\subsection{The Robot-Native Regime}
By robot-native supervision, we mean physical experience that is already represented in the coordinate system of a robot-learning problem. In the standard case, this consists of trajectories collected from a particular embodiment, where robot observations are paired with robot actions and, in some cases, language instructions, task labels, rewards, or success indicators. The observations may include camera images, proprioceptive states, end-effector poses, tactile readings, force-torque measurements, or other sensor streams; the actions may correspond to joint commands, end-effector displacements, gripper states, velocity commands, or higher-level skill primitives. This form of data is powerful because it directly matches the supervised or reinforcement-learning objective: a model can imitate the demonstrated action, optimise the provided reward, or condition behaviour on the associated task description. In other words, the data is useful because it has already been grounded into a particular robot body, action space, sensor suite, and task definition.

This regime has enabled a series of increasingly ambitious efforts to scale robot learning through larger and more diverse robot datasets. Early multi-robot datasets such as RoboNet demonstrated the value of sharing robotic experience across platforms, containing 15 million video frames from seven robot platforms and supporting both video-prediction and inverse-model learning \citep{RoboNet}. More recent datasets have expanded the scale and diversity of robot-native supervision: BridgeData V2 \citep{walke2023bridgedata} provides roughly sixty thousand manipulation trajectories across twenty-four environments on a low-cost robot platform, while DROID contains approximately seventy-six thousand demonstration trajectories, or 350 hours of interaction data, collected across hundreds of scenes and dozens of tasks by geographically distributed data collectors \citep{khazatsky2024droid}. RH20T further broadens this trend by collecting over 110,000 contact-rich manipulation sequences with visual, force, audio, and action information, together with corresponding human demonstration videos, making it especially relevant for studying multimodal grounding in manipulation \citep{fang2023rh20t}. These datasets make clear that robot-learning performance improves when models are exposed to more tasks, objects, environments, and embodiments. They also show that data diversity is not a detail but a central requirement for generalisation beyond a single laboratory setup.

In parallel, robot-native trajectories have become the substrate for increasingly generalist robot policies. BC-Z studied how scaling real-robot imitation data across more than one hundred tasks can enable zero-shot generalisation to unseen manipulation tasks, including conditioning policies on language or videos of humans performing the task \citep{BCZ}. RT-1 showed that a transformer policy trained on approximately 130,000 real-robot episodes collected across thirteen robots and more than 700 tasks could produce broad language-conditioned manipulation behaviour \citep{brohan2022rt}. RT-2 extended this direction by co-training vision-language models on web-scale vision-language data and robot trajectories, representing robot actions as tokens so that semantic knowledge from internet-scale pretraining could be transferred into robotic control \citep{zitkovich2023rt}. SayCan \citep{ahn2022can} and PaLM-E \citep{driess2023palm} similarly illustrate how large language or multimodal models can contribute to semantic reasoning and planning, but still require grounding through robot affordances, embodied observations, or learned low-level skills.

The same trend appears in cross-embodiment and open-source generalist policy efforts. Open X-Embodiment and RT-X pooled more than one million real robot trajectories from 22 robot embodiments by aggregating datasets from many research laboratories into a common format, making cross-embodiment training a practical research direction \citep{o2024open}. Octo then showed that an open-source generalist policy can be pretrained on 800,000 trajectories from Open X-Embodiment and adapted to new observation and action spaces \citep{team2024octo}, while RoboCat \citep{bousmalis2023robocat} explored a self-improving generalist manipulation agent trained on action-labelled experience across multiple robots and tasks. Systems such as Dobb-E further push robot learning into less curated settings by collecting household demonstrations and adapting policies for new home tasks, highlighting both the promise and the messiness of real-world deployment \citep{shafiullah2023bringing}. Taken together, these works mark a major shift from single-task, single-robot learning toward broader robot foundation models. Yet they also reinforce the central premise of this section: the most effective supervision still largely arrives as robot-grounded trajectories, where actions, observations, task descriptions, and success signals have already been made legible to a robot-learning algorithm.

A complementary line of work has improved the policy-learning machinery applied to robot-native trajectories. Diffusion Policy, for example, formulates visuomotor control as conditional denoising over action sequences, showing that diffusion models can represent multimodal action distributions and produce strong manipulation policies from demonstration data \citep{chi2025diffusion}. ALOHA \citep{fu2024mobile, zhao2024aloha} and Action Chunking with Transformers \citep{george2023one, bharadhwaj2024roboagent, zhang2025act} similarly show that carefully designed low-cost teleoperation systems, combined with sequence-level imitation learning, can acquire fine-grained bimanual manipulation skills from real-world demonstrations

Those developments have contributed to a rapid expansion of vision-language-action models, i.e., models that map observations and textual descriptions to robotic actions. Early systems such as Gato \citep{reed2022generalist} helped establish the idea that a single transformer-style agent could operate across modalities and domains, including real robot control. At the same time, SayCan \citep{ahn2022can} and PaLM-E \citep{driess2023palm} showed how large language or multimodal models could support embodied reasoning, planning, and affordance-aware skill selection, provided they were grounded through robot skills, observations, or low-level policies. RT-2 then made the VLA framing more explicit by co-training web-scale vision-language models with robot trajectories and representing robot actions as tokens, allowing semantic knowledge from internet-scale pretraining to be transferred into robotic control \citep{zitkovich2023rt}. Related systems explored different forms of multimodal task specification and VLM adaptation: VIMA formulates robot manipulation as multimodal prompting over interleaved text and visual tokens, while RoboFlamingo adapts open vision-language foundation models for language-conditioned robotic manipulation \citep{jiang2023vima}. Moreover, OpenVLA is a representative open-source example: it trains a 7B-parameter VLA on approximately 970,000 real-world robot demonstrations from Open X-Embodiment, taking images and language instructions as input and producing robot actions as output \citep{kim2024openvla}. Physical Intelligence’s $\pi_0$ similarly frames generalist robot control as a vision-language-action problem, using a flow-matching architecture built on top of a pretrained vision-language model to inherit internet-scale semantic knowledge while producing continuous robot actions \citep{black2024pi_0}. A related line of work focuses on the action-generation mechanism itself: CogACT separates the vision-language reasoning component from a specialised action module and studies diffusion action transformers for action-sequence modelling, while RoboMamba explores state-space-model architectures for efficient vision-language-action reasoning and manipulation \citep{li2024cogact,liu2024robomamba}. Furthermore, FAST and related action-tokenisation methods address a complementary bottleneck: how high-frequency continuous robot actions should be compressed or tokenised so that they can be modelled efficiently by VLA architectures \citep{moodley2024multi, zhong2025survey,pertsch2025fast, liu2025faster, dong2026actioncodec}.

Other recent models have focused on more specialised ingredients needed for scalable VLA control: SpatialVLA incorporates explicit spatial representations for robot manipulation and is trained on approximately 1.1 million real robot episodes, while RDT-1B \citep{liu2024rdt} uses a diffusion-transformer architecture for bimanual manipulation and is pretrained on more than one million multi-robot episodes \citep{zhang2024vla, qu2025spatialvla, patratskiy2025spatial, feng2025spatial}. This spatial focus is also reflected in 3D-VLA, which argues that robot foundation models should move beyond 2D visual inputs by linking 3D perception, embodied reasoning, action prediction, and generative world modelling \citep{qu2025spatialvlaexploringspatialrepresentations}. Related spatially grounded VLA efforts include 3DS-VLA, which uses 3D spatial constraints and point-cloud information for manipulation \citep{li20253ds}; GeoVLA, which integrates depth-derived point clouds through a point encoder and a 3D-enhanced action expert \citep{sun2025geovla}; GraphCoT-VLA, which introduces a 3D pose-object graph and structured chain-of-thought reasoning for ambiguous manipulation instructions \citep{huang2026graphcot}; and Avi, which reframes robotic action generation as a problem of 3D perception and language-grounded spatial reasoning rather than only low-level policy learning \citep{song2025avi}. Another emerging concept for endowing robot policies with strong spatial-physical priors is to build policies on top of video generation backbones.  One such paradigm is the World -Action Model (WAM), which predicts video frames as well as action chunks.  DreamZero \cite{ye2026world} and Unified Video Action (UVA) Model \cite{li2025unified} are both prime examples. 

More recent humanoid-focused systems extend the VLA paradigm from tabletop manipulation toward whole-body and dexterous control. NVIDIA Isaac GR00T N1 explicitly targets generalist humanoid robots and is described as a dual-system VLA: a vision-language module interprets the scene and instruction, while a diffusion-transformer action module generates real-time motor commands for humanoid control \citep{bjorck2025gr00t}. Importantly, GR00T N1 is trained on a mixture of egocentric human videos, real and simulated robot trajectories, and synthetic data, making it a useful example of the broader data mixture now being explored for humanoid robotics. Gemini Robotics follows a similar foundation-model direction, presenting a generalist VLA built on Gemini that can directly control robots from visual observations and language instructions, with later on-device variants reported to support deployment without constant internet connectivity and adaptation to other platforms such as Apptronik’s Apollo and Franka FR3 \citep{team2025gemini}. Figure’s Helix is another humanoid-focused VLA system, presented as a model that unifies perception, language understanding, and learned control for full upper-body humanoid manipulation, including wrists, torso, head, and fingers \citep{figure2025helix}. Beyond these frontier model releases, recent research has begun to study humanoid-specific VLA structure more directly. LeVERB proposes a hierarchical VLA framework for humanoid whole-body control, learning a latent action vocabulary from synthetically rendered kinematic demonstrations and using a reinforcement-learned whole-body controller to execute dynamics-level commands \citep{xue2025leverb}. WholeBodyVLA similarly targets closed-loop humanoid loco-manipulation, learning unified latent actions from action-free egocentric videos and combining them with a locomotion-oriented control policy \citep{jiang2025wholebodyvla}. HuMI addresses the data-collection bottleneck from a different angle, using portable robot-free demonstrations to learn whole-body humanoid manipulation skills and reporting improved data-collection efficiency compared with teleoperation \citep{nai2026humanoidmanipulationinterfacehumanoid}. HEX focuses on cross-embodiment whole-body manipulation, introducing humanoid-aligned state representations and expert modules to improve coordinated control on full-sized bipedal robots \citep{bai2026hex}. Industry systems such as Skild Brain also frame humanoid and general-purpose robot control as an omni-bodied foundation-model problem trained from simulation, human action videos, and real-world robot feedback \citep{skild2025brain}.

The breadth of this literature shows how far robot learning has moved beyond single-task, single-platform imitation. Yet the underlying supervision regime remains largely robot-native. The strongest results still depend on experience that has already been expressed as observations paired with actions, tasks, rewards, demonstrations, or success labels. This is precisely why these systems work, but it also identifies the scaling bottleneck: most of the physical behaviour available in the world does not arrive with explicit robot actions.

\underline{\textbf{Takeaway.}} Robot-native supervision has enabled the most impressive progress in generalist robot policies. However, its strength is also its limitation: the data has already been expressed in the coordinate system of robot learning. Actions, task labels, embodiment constraints, and success signals are either collected directly or curated afterwards. This makes VLA scaling powerful, but still dependent on supervision that has already been grounded.

\subsection{Learning from Weakly Grounded Physical Observations}
To enable \emph{true scalability}, a growing body of work asks whether robot learning can benefit from physical observation that is not natively action-labelled. This direction is motivated by a simple asymmetry: robot action-labelled trajectories are expensive to collect, but videos of humans and physical interactions are abundant. Human videos contain information about object affordances, task structure, contact events, temporal progress, and failure recovery, even when they do not specify the motor commands that a robot should execute. The central question is whether such passive physical observations can be converted into useful learning signals for robot policies.

A useful way to formalise the problem is to distinguish between observed physical change and executable robot action. A video provides an observation sequence: $\textbf{o}_{1:\mathsf{T}} = \langle o_1, \dots, o_\mathsf{T} \rangle$, but not the robot action sequence: $\textbf{a}_{1:\mathsf{T}}=\langle a_1, \dots, a_{\mathsf{T}}\rangle$. Robot-native imitation learning assumes access to pairs $(o_t, a_t)$, while learning from human or internet videos usually provides only $\textbf{o}_{1:\mathsf{T}}$, sometimes with language $\texttt{L}_{1:\mathsf{T}}$, captions, or weak task metadata. The latent variables are therefore action-like representations $\textbf{z}_{1:\mathsf{T}}$ that explain transitions from $o_t$ to $o_{t+1}$ such that: $z_{t} \sim q(\cdot|o_{t}, o_{t+1}, \texttt{L}_{t}, \texttt{L}_{t+1})$. Generally, those latent variables are not yet tied to any particular robot embodiment. The hope is that such latent actions capture task-relevant changes in the world, e.g., moving, grasping, opening, placing, inserting, aligning, and can later be mapped to embodiment-specific robot actions. This makes latent-action learning a natural bridge between passive video and robot-native control. 

\paragraph{Representation Learners.} Several representation-learning works can be viewed as indirect steps in this direction. R3M pretrains visual representations on the Ego4D human video dataset using time-contrastive learning, video-language alignment, and sparsity regularisation, then uses the frozen representation for downstream robot manipulation policies \citep{nair2022r3m}. VIP learns visual representations from human videos by using temporal distance as a proxy for task progress, producing features that can support robotic reinforcement learning and imitation \citep{ma2022vip}. MVP studies masked visual pretraining for robot manipulation, while VC-1 evaluates a range of pretrained visual representations across robotic tasks and argues for large-scale visual pretraining as a reusable perceptual substrate for embodied control \citep{radosavovic2023real, majumdar2023we}. These methods do not directly recover robot actions from video, but they show that passive human or internet-scale visual experience can improve the perceptual and task-relevant features used by robot policies. R3M, in particular, explicitly studies how human video pretraining can enable data-efficient manipulation learning. However, these representation-learning methods mostly do not fully resolve the correspondence problem: how an observed physical change in human or internet video becomes a reward, subgoal, latent action, or executable control signal for a particular robot.


Earlier work on imitation from observation and cross-embodiment learning exposed the same weak-grounding problem in a more explicit form. Time-Contrastive Networks learned viewpoint-invariant representations from unlabeled multi-view videos and used the resulting embeddings for robotic imitation and reward learning, including imitation from human demonstrations without explicit action correspondence \citep{sermanet2018timecontrastivenetworksselfsupervisedlearning}. AVID addressed the human-to-robot appearance gap by translating human demonstration videos into robot-domain visual instructions, which were then used as rewards for model-based reinforcement learning \citep{smith2020avidlearningmultistagetasks}. XIRL formulated the problem as cross-embodiment inverse reinforcement learning, learning vision-based rewards from videos of agents with different bodies, actions, and end-effector dynamics \citep{zakka2021xirlcrossembodimentinversereinforcement}. Similarly, DVD learned generalisable reward functions from a mixture of in-the-wild human videos and a small amount of robot data, using functional similarity between human and robot behaviour as the supervision signal \citep{chen2021learninggeneralizableroboticreward}. Together, these works show that the core obstacle is not merely the absence of robot actions, but the broader correspondence problem: how to align observed physical progress with what a different robot body can perceive, value, and execute.


\paragraph{Latent-Action Approaches.} More recent latent-action approaches attack the missing-action-label problem more directly. Latent Action Pretraining (LAPA) proposes an unsupervised method for pretraining VLA models without ground-truth robot action labels \citep{ye2024latent}. It first learns a discrete latent action space from video transitions using a VQ-VAE-style objective, then trains a latent VLA model to predict these latent actions from observations and task descriptions, before fine-tuning on smaller robot datasets to map latent actions to executable robot actions. This is particularly relevant to our thesis because LAPA treats internet-scale video not merely as semantic or perceptual pretraining data, but as a source of action-like structure. In other words, it takes the first steps to ask whether the physical changes visible in the video can be compressed into a reusable action vocabulary before being grounded into a particular robot body. A related line of work learns action-like abstractions from heterogeneous embodiments and viewpoints. UniVLA, for example, derives task-centric latent actions in an unsupervised manner, allowing the model to leverage data from arbitrary embodiments and camera perspectives without requiring action labels \citep{bu2025univla}. Strictly speaking, however, such latent variables are better understood as transition codes or physical-change descriptors until they are grounded in a specific robot embodiment. They become robot actions only when an embodiment-conditioned decoder can map them to commands that, when executed, reproduce the intended physical change. 

\paragraph{Task-Progress Signals.} Another branch uses video-language models to infer task progress, rewards, and success from passive observation. Rather than asking videos to provide actions, these methods ask videos to provide supervision over \emph{what matters}. PROGRESSOR learns a task-agnostic reward function from unlabelled videos and uses self-supervised refinement to provide dense rewards for goal-conditioned policy learning \citep{ayalew2025progressor}. Adapt2Reward transfers video-language models into language-conditioned reward functions using limited robot video data \citep{yang2024adapt2reward}. ReWiND trains a reward model to predict video progress rewards from image or rollout embeddings and language instructions, using video rewinding and misaligned video-language pairs as negative supervision \citep{zhang2025rewind}. TimeRewarder derives progress signals from passive videos, including both robot demonstrations and human videos, by modelling temporal distances between frame pairs \citep{liu2025timerewarder}. Stage-Aware Reward Model (SARM) uses dense subtask labels to help supervised a reward model to determine fine-grained task progress \cite{chen2025sarm}.  These works are crucial because they suggest that passive video may be useful not only for perceptual pretraining or latent actions, but also for grounding task progress and reward.

This literature reveals a useful taxonomy of weak physical supervision. Passive videos can provide at least four kinds of signal. First, they can provide visual representations, as in R3M, VIP, MVP, and VC-1. Second, they can provide latent action codes, as in LAPA and UniVLA. Third, they can provide task-progress and reward signals, as in PROGRESSOR, Adapt2Reward, ReWiND, TimeRewarder, and SARM. Fourth, they can provide behavioural priors about object use, affordances, contact, and temporal task structure. Recent surveys on learning from human or internet video emphasise this same opportunity while highlighting the unresolved challenges: videos lack robot actions, viewpoints and embodiments differ, physical contact and forces are often unobserved, and human strategies may not be directly executable by a robot \citep{eze2025learningwatchingreviewvideobased, feng2026human}.

However, weak physical supervision does not remove the need for grounding; it relocates it. A latent action learned from videos is not yet a robot command. A progress signal inferred from temporal order is not necessarily a reward for a new embodiment. A visual representation trained on human videos may encode objects and affordances, but not the contact dynamics or force constraints needed for manipulation. Thus, the central challenge is not simply to pretrain on more video. It is to determine which variables should be extracted from video, how those variables should be grounded into robot morphology and control, and how errors in this grounding affect downstream policy learning. In this sense, learning from action-free video is one of the clearest examples of the broader thesis of this paper: the world contains abundant physical experience, but robotics still lacks reliable mechanisms for transforming that experience into robot-usable supervision.

\underline{\textbf{Takeaway.}} Passive video can provide representations, progress signals, latent actions, and behavioural priors. But these signals are not yet robot supervision. A latent action is not a command; a progress signal is not necessarily a reward; and a human strategy may not be executable by a robot. Video expands the source of physical experience, but it also makes the grounding problem unavoidable.

\subsection{Generating Physical Experience }
A second response to the cost of robot-native supervision is not to infer labels from existing observations, but to generate additional physical experience. If robot trajectories are expensive because they must be enacted on hardware, then simulation, synthetic demonstration generation, and learned world models offer a complementary route: they can expose policies to more tasks, objects, initial conditions, failures, and counterfactual outcomes than would be practical to collect directly in the real world. This shifts the data-scaling question from “How many robot trajectories can we collect?” to “How faithfully can we generate experience that preserves the physical structure needed for control?” In this view, simulation and world models are not merely data factories; they are mechanisms for expanding the reachable distribution of physical interactions. Their value depends on whether the generated experience captures the quantities that matter for downstream learning: geometry, object state, contact, dynamics, embodiment constraints, task progress, and success or failure.

\paragraph{The Simulation Route.} A first route to generating physical experience is simulation. Simulation environments and benchmarks provide controllable worlds in which tasks, objects, layouts, initial states, and demonstrations can be generated at a scale that would be difficult to reproduce on hardware. RLBench is an early and influential example, providing 100 hand-designed manipulation tasks, multimodal observations, language descriptions, and an effectively unlimited supply of demonstrations generated through motion planning \citep{james2019rlbenchrobotlearningbenchmark}. Meta-World \citep{yu2020meta} similarly helped standardise multi-task and meta-reinforcement learning for robotic manipulation, while ManiSkill focuses on generalisable manipulation from 3D visual inputs in a full-physics simulator with diverse object geometries \citep{mu2021maniskillgeneralizablemanipulationskill}. CALVIN extends this direction toward long-horizon language-conditioned manipulation, asking agents to compose many behaviours from language instructions in simulated environments \citep{mees2022calvinbenchmarklanguageconditionedpolicy}. LIBERO further studies lifelong robot learning, knowledge transfer, and task-ordering effects across 130 language-conditioned manipulation tasks \citep{liu2023liberobenchmarkingknowledgetransfer}. Together, these environments have been crucial because they make robot learning reproducible, scalable, and comparable across methods. Yet they also illustrate a recurring assumption: the simulator designer has already specified the relevant state, action space, task definitions, object assets, and success conditions.

\paragraph{The Data Generation Route.} A second route is to use simulation not only as an evaluation environment, but as a data-generation engine. MimicGen is a particularly important example: it automatically synthesises large-scale demonstration datasets in a simulator from a small number of human demonstrations by adapting demonstration segments to new object poses and contexts, generating more than 50,000 demonstrations from fewer than 200 seed demonstrations across 18 tasks \citep{mandlekar2023mimicgendatagenerationscalable}. RoboCasa scales this idea toward everyday household manipulation by providing a large-scale simulated kitchen environment for training generalist robots, and its authors report clear scaling trends from using synthetically generated robot data for imitation learning \citep{nasiriany2024robocasa}. RoboCasa365 pushes this further with 365 everyday household tasks, 2,500 kitchen scenes, and over 2,000 hours of robot interaction data, including both human demonstrations and synthetic demonstrations generated with MimicGen \citep{nasiriany2026robocasa365}. RoboGen takes a more automated route, using foundation and generative models to construct tasks, scenes, and training data in simulation so that robots can acquire diverse skills at scale \citep{wang2024robogenunleashinginfinitedata}. These systems are important because they shift robot data collection from “manually collect every trajectory” to “collect a small amount of seed experience, then generate variations.” The open question is whether those variations preserve the physical details that matter for real control, especially contacts, object stability, friction, deformation, and failure modes. 

\textbf{The Real-to-Sim-to-Real Route.} Instead of relying on a hand-built simulator, these methods attempt to reconstruct or approximate a real environment, use simulation to expand or robustify learning, and then return the resulting policy to the physical world. RialTo is a representative example: it constructs digital-twin simulation environments from small amounts of real-world data and uses reinforcement learning in simulation to robustify imitation policies before real-world deployment \citep{torne2024reconcilingrealitysimulationrealtosimtoreal}. More recent work has begun to use 3D Gaussian Splatting and related reconstruction methods to make digital twins more visually faithful and easier to build from real observations. RL-GSBridge introduces a 3D-Gaussian-Splatting-based real-to-sim-to-real reinforcement-learning framework for robotic manipulation, using reconstructed scenes to support zero-shot sim-to-real transfer for vision-based control \citep{wu2025rlgsbridge3dgaussiansplatting}. Real-is-Sim similarly uses a dynamic digital twin based on Embodied Gaussians throughout data collection, training, policy evaluation, and deployment, aiming to reduce the gap between offline policy evaluation and real-world success \citep{abouchakra2025realissimbridgingsimtorealgap}. Related real-to-sim evaluation work constructs soft-body digital twins from real-world videos and renders robots, objects, and environments with photorealistic fidelity using 3D Gaussian Splatting, showing that simulated rollouts can correlate with real policy performance on deformable manipulation tasks such as plush packing and rope routing \citep{zhang2025realtosimrobotpolicyevaluation}. RoboGSim also follows this direction, presenting an interactive real2sim2real Gaussian-splatting platform for demonstration synthesis, novel-scene and novel-object data scaling, and closed-loop policy evaluation \citep{li2025robogsimreal2sim2realroboticgaussian}. For robot navigation tasks in particular, policies trained in 3D Gaussian Splatting simulators have been shown to transfer zero-shot to real world deployment.  SOUS VIDE and SINGER are examples of such approaches using an imitation learning paradigm \cite{low2025sous,adang2025singer}, while GRaD-Nav and GRaD-Nav++ train with RL and exploit end-to-end differentiability of 3DGS rendering \cite{chen2025grad,chen2025grad++}.

This direction builds on a broader sim-to-real literature that has long studied how generated experience can transfer to physical robots despite imperfect simulators. Domain randomisation is one of the dominant strategies: rather than fitting a single simulator, it trains policies over a distribution of simulated dynamics or visual parameters, so that the real world appears as a single variation within the training distribution \citep{muratore2022robotlearningrandomizedsimulations}. Those ideas have also been used in legged robotics: learning agile and dynamic motor skills for ANYmal showed that neural policies trained in simulation can transfer to a real quadruped, while later massively parallel reinforcement-learning work showed that locomotion policies can be trained in minutes in simulation and transferred to real robots \citep{Hwangbo_2019, rudin2022learningwalkminutesusing}. Domain randomisation in simulation can also be combined with online latent parameter adaptation to better facilitate the sim-to-real transfer, for example, using the Rapid Motor Adaptation (RMA) approach for legged locomotion \cite{kumar2021rma}, and further adapted to manipulation tasks in \cite{wang2025phys2real}.  These works are not always “real-to-sim-to-real” in this sense. Still, they are essential context because they show why simulation became attractive in the first place: it generates experience cheaply, safely, and at scale, but only transfers when the simulator captures or randomises the physical factors that matter.

\textbf{The World-Modelling Route.} A fourth route is to replace or augment explicit simulation with learned world models. The idea has a long history: rather than learning only a reactive policy, an agent can learn an internal predictive model of how the environment changes under its actions and then use that model for planning, imagination, or policy improvement. Schmidhuber’s early work on “making the world differentiable” already proposed reinforcement learning and planning with recurrent neural networks, including a controller and a learned world model that predicts environmental dynamics \footnote{\url{https://people.idsia.ch/~juergen/world-models-planning-curiosity-fki-1990.html}.}. This view was later popularised in deep learning by Ha and Schmidhuber’s World Models, which trained a generative recurrent model of the environment and then learned compact policies from the model’s latent representations, including policies trained inside the model’s own “dreamed” rollouts before being transferred back to the real environment \citep{Jurg, ha2018recurrentworldmodelsfacilitate}. In modern reinforcement learning, PlaNet and Dreamer made this idea practical at scale by learning compact latent dynamics from pixels and improving policies through imagined futures; DreamerV3 later showed that a single world-model algorithm can solve a wide range of control tasks with fixed hyperparameters, while DayDreamer demonstrated that Dreamer-style world models can be learned directly on physical robots for locomotion, manipulation, and navigation without relying on a hand-built simulation \citep{Dreamerv1, hafner2019planet, DreamerV2,wu2022daydreamerworldmodelsphysical, hafner2025mastering}. 

In robotics, this tradition is being extended from latent dynamics for control toward models that can generate or predict physically meaningful experience. RoboDreamer learns compositional video world models for robot imagination by factorising video generation according to language-derived primitives, allowing it to synthesise plans for unseen combinations of objects and actions \citep{zhou2024robodreamerlearningcompositionalworld}. UniSim asks whether a universal interactive simulator can be learned from diverse datasets, combining information from images, robotics data, and navigation data to simulate the visual outcomes of high-level instructions and low-level controls; it also shows that policies trained in the learned simulator can transfer to real-world deployment in some settings \citep{yang2024learninginteractiverealworldsimulators}. DeepMind’s Genie extends this line toward generative interactive environments trained from unlabelled internet videos, introducing a spatiotemporal tokeniser, an autoregressive dynamics model, and a learned latent action model that enables frame-by-frame control without ground-truth action labels \citep{bruce2024genie}. These systems are exciting because they blur the boundary between video generation, simulation, and robot learning: generated experience is no longer just passive video, but potentially an action-conditioned environment in which agents can plan or train.

However, robotics places stricter demands on world models than visual plausibility alone. A useful robot world model must preserve the variables that matter for action: 3D geometry, object permanence, contact, material properties, constraints, forces, and the consequences of robot motion. This has motivated a growing line of \emph{3D, object-centric, and physics-grounded world models}. Object-centric world models such as FOCUS argue that manipulation requires representing objects and their interactions explicitly, enabling more efficient exploration of robot-object dynamics \citep{ferraro2023focusobjectcentricworldmodels}. Related language-guided object-centric world models predict future states in compact object-centric representation spaces rather than only generating pixels, improving efficiency for visuo-linguo-motor control \citep{jeong2025objectcentricworldmodellanguageguided}. PointWorld scales this idea to 3D by unifying state and action in a shared spatial domain and predicting full-scene 3D point flow from RGB-D observations and robot actions \citep{huang2026pointworldscaling3dworld}. ParticleFormer similarly learns a transformer-based 3D point-cloud world model for multi-object, multi-material manipulation, predicting dynamics directly from real-world robot perception data and supporting downstream manipulation through model-predictive control \citep{huang2025particleformer3dpointcloud}. These methods are especially relevant because they move world modelling away from generic image prediction and toward spatially grounded, action-conditioned dynamics. 

When developing world models for robot imagination, planning, data generation, or evaluation, it is critical to assess the confidence of a world model prediction.  Learned world models, as all learned models, are subject to errors and hallucinations when queried outside the distribution of their training data.  For a robot using a world model for planning, this can lead to a viscous cycle, where a hallucination leads to a poor control choice, leading the world model further away from its training distribution invoking further hallucinations, which further degenerate control effectiveness.  Therefore, for robotics in particular, world models need to have a calibrated estimate of the certainty of their own predictions. Early work on this topic includes  \cite{mei2025world}, which learns a latent uncertainty quantification with a VAE approach, statistically calibrated to yield interpretable uncertainties. Latent uncertainties can then be passed to the pixel level for visualisation of uncertain regions of the predicted image.  The authors of \cite{li2025uncertainty} further show the importance of uncertainty quantification for a world model used as an environment for training reinforcement learning policies.  In \cite{ward2026foundational} the authors show a world model with calibrated uncertainty in the latent space can be used to detect runtime errors in a VLA manipulation policy.  Uncertainty quantification for world models is still a young area, but we anticipate work on this topic to grow as world models become more integrated into robot autonomy stacks.

Another way to prevent erroneous world model predictions is to structure a world model as a combination of a neural scene representation with physical simulation.  This can be viewed as a modern, 3D neural-scene version of a much older model-based robot-learning idea: learn or construct a predictive model of the world, then use it for planning, policy optimisation, or data-efficient control. Classical robot-learning work already emphasised this model-based view \citep{Kober2013IJRR} and discussed the importance of models, data efficiency, exploration, and safe real-world learning in robotics. PILCO \citep{deisenroth2011pilco} and its variants \citep{polymenakos2019safe, cowen2022samba} are another canonical example, using probabilistic Gaussian-process dynamics models to perform data-efficient policy search with uncertainty-aware long-term predictions.

A more structured version of this model-based view attempts to ground learned dynamics directly in physical laws or relational structure, rather than treating the transition model as an unconstrained black box. Deep Lagrangian Networks impose the structure of Lagrangian mechanics on neural models of robot dynamics, learning physically plausible inertia, Coriolis, gravitational, and control-dependent terms for model-based control \citep{janDeepLagrange}. Hamiltonian Neural Networks learn a Hamiltonian function and use Hamilton’s equations to produce energy-aware dynamics \citep{HNNs}, while Lagrangian Neural Networks parameterise a Lagrangian and derive equations of motion through the Euler-Lagrange equations \citep{LNNs}. Related geometric dynamics models, such as Symplectic ODE-Net, enforce Hamiltonian or symplectic structure to improve long-horizon physical prediction \citep{SODEs}. A complementary family represents the world as interacting objects, particles, or relations: Interaction Networks introduced graph-based reasoning over objects and relations for physical prediction \citep{Inter}, Neural Physics Engines model physical systems through learned object-centric interactions \citep{chang2017compositionalobjectbasedapproachlearning}, and Graph Networks as Learnable Physics Engines showed that graph networks can simulate complex physical systems and generalise across different numbers and configurations of objects \citep{sanchezgonzalez2018graphnetworkslearnablephysics}. Later graph-network simulators scaled this idea to particle-based fluids, rigid materials, and deformable systems \citep{sanchezgonzalez2020learningsimulatecomplexphysics}. These works are important because they show that a robot world model can be grounded not only by more data, but also by inductive biases that encode energy, geometry, object relations, constraints, and interaction structure. 

Of course, the contemporary versions of those ideas increasingly use learned visual and geometric representations as the substrate of the world model. LeCun’s “A Path Towards Autonomous Machine Intelligence” argues that autonomous agents need predictive world models for planning under uncertainty, and the JEPA family operationalises this vision by predicting in abstract representation spaces rather than directly reconstructing pixels \citep{lecun2022path}. I-JEPA introduced image-based joint-embedding predictive learning, while V-JEPA extended this to video by predicting masked parts of videos in representation space rather than pixel space \citep{assran2023self}. V-JEPA 2 is especially relevant here because it combines internet-scale video with a small amount of robot interaction data and reports prediction, planning, and zero-shot robot control capabilities, making it one of the clearest recent links between JEPA-style world models and embodied control \citep{assran2025vjepa2selfsupervisedvideo}. 

A further recent strand combines neural scene representations with physical simulation, aiming to build world models that are both visually grounded and physically actionable. Physically Embodied Gaussian Splatting proposes a dual Gaussian-particle representation that couples visual rendering with particle-based physical prediction and online correction based on observations, enabling the model to reason about present and future physical states while staying synchronised with the real world \citep{xie2024physgaussianphysicsintegrated3dgaussians}. Gaussian World Models similarly use 3D Gaussian Splatting as a dynamic world representation for robotic manipulation, supporting action-conditioned 3D video prediction, imitation-learning representations, and model-based reinforcement learning \citep{lu2025gwm}. ContactGaussian-WM pushes this direction toward contact-rich manipulation by learning a physics-grounded rigid-body world model from sparse contact-rich videos, combining a unified Gaussian representation for visual appearance and collision geometry with differentiable contact dynamics and closed-form physical reasoning; it reports applications to data synthesis and real-time model-predictive control \citep{wang2026contactgaussianwmlearningphysicsgroundedworld}. Physics-informed world models for non-prehensile or deformable manipulation similarly attempt to inject physical structure into learned prediction rather than relying on unconstrained video generation, for example, by combining differentiable physics, visual observations, and physics-aware randomisation for robust sim-to-real manipulation \citep{pinwm2025}. These approaches make explicit what many video world models leave implicit: for robot planning, the generated future must obey enough geometry, contact, dynamics, and physical constraints to be useful for control.

Taken together, these strands suggest that world models for robotics should not be evaluated only by how realistic their generated observations appear. Their central purpose is to make physical experience counterfactual. A robot should be able to ask what would happen if it pushed at a different point, grasped with a different orientation, opened a drawer further, inserted an object at a slightly different angle, or stopped applying force. In a robot-native dataset, only the executed trajectory is observed; in a world model, the agent can imagine alternatives. But this advantage is meaningful only if the imagined futures preserve the variables that determine success and failure: object state, geometry, contact, force, stability, material response, embodiment constraints, and task progress. A visually plausible prediction that ignores contact, mass, friction, or physical feasibility may help representation learning, but it is not yet a reliable substrate for robot control. Thus, learned world models sharpen rather than solve the grounding problem: they promise scalable generated experience, but only if the generated experience is physically grounded and actionable.

\underline{\textbf{Takeaway.}} Generated experience is useful only when it preserves the physical variables that determine control. A visually plausible rollout that ignores contact, force, friction, or stability is not yet reliable robot supervision. The value of simulation and world models is therefore not visual realism alone, but physically grounded counterfactual experience.

\section{The Missing Components for Physical Intelligence} 
The survey above suggests that the next step in robotics is not simply to train larger policies, collect more demonstrations, or build more visually realistic simulators. These directions are necessary, but \emph{incomplete.} What is missing is a set of components that transform broad physical experience into grounded, deployable robot intelligence. Today, robot learning often begins once the relevant variables have already been specified: observations, actions, task labels, rewards, success metrics, and embodiment constraints. Future physical-intelligence systems must instead recover these variables from weaker, messier, and more diverse sources of experience: human motion, internet video, wearable sensing, tactile streams, robot rollouts, simulation, language, and failure traces. In this sense, the central challenge is: how should a robot-learning system be organised so that the world itself becomes a source of supervision? 
\begin{figure}
    \centering
\includegraphics[width=\linewidth]{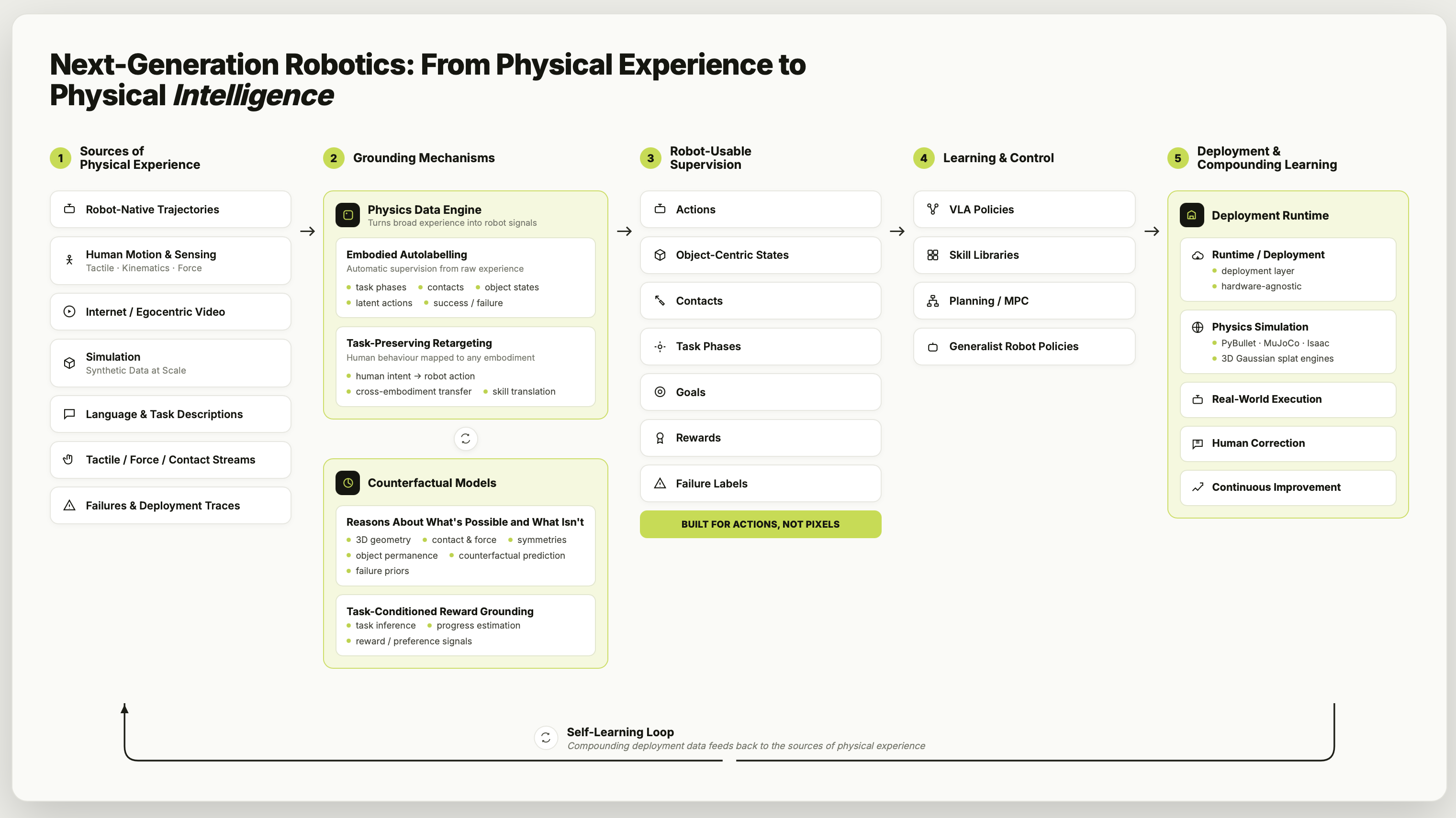}
    \caption{Next generation robotics will come from advances that go well beyond scaling vision language action (VLA) models. }
    \label{fig:placeholder}
\end{figure}
We argue that such a system requires four missing components. First, it needs a physical data engine that can ingest heterogeneous experience and convert it into structured signals such as object states, contact events, task phases, latent actions, and success or failure labels. Second, it needs task-preserving retargeting, so that human behaviour and inferred skills can be translated into executable actions across different robot embodiments. Third, it needs physics-grounded world models that predict not only plausible future observations, but the physical consequences of action: geometry, contact, force, stability, constraints, and material response. Fourth, it needs a self-improving deployment loop, where each robot deployment produces new data, new failures, and new corrections that compound into broader competence rather than remaining isolated engineering effort. The proposed stack should therefore be understood as closed-loop rather than purely feed-forward. Rather than treating deployment only as evaluation, a physical-intelligence system should turn real rollouts, failures, and human corrections into structured supervision. Once grounded into contacts, object-state changes, reward errors, or failure modes, these traces can update the policy, reward model, world model, or retargeting model, allowing robots to improve on the actual tasks they face.

\emph{We believe that the next foundation model for robotics will not be only a VLA or a world-model. It will be a pipeline that grounds physical experience into actions, rewards, world models, and deployment feedback.}

\subsection{Physical Data Engines and Embodied Autolabelling}
The first missing component is a physical data engine: a system that turns heterogeneous physical experience into structured learning signals for robots. Today, much of robot learning begins after data has already been made convenient for the policy: observations are paired with robot actions, demonstrations are segmented into tasks, success conditions are specified, and rewards are either hand-designed or manually labelled. This assumption is powerful, but it does not scale to the broader physical world. Human motion, internet video, wearable sensing, tactile streams, factory workflows, failures, and deployment traces all contain useful information about physical interaction, but they do not arrive as clean robot-training examples. A physical data engine is the missing layer that converts this messy experience into the variables that robot learning actually requires: object states, contacts, task phases, latent actions, goals, rewards, and success or failure labels.

The key idea is that physical experience should not be treated as raw video or unstructured logs. It should be treated as partially labelled interaction data. A person performing a task reveals more than pixels: their motion indicates intent, their hands reveal contact, object movement reveals causal structure, pauses and corrections reveal uncertainty, and the final configuration reveals something about success. Similarly, a failed robot rollout is not merely a bad trajectory; it is evidence about what the policy misunderstood, which contact was missed, which object state was unstable, or which subgoal was not achieved. Even a failure is an opportunity to learn a new skill, as long as it is properly labeled.  Perhaps an object was mistakenly dropped due to a poor grasp. In a future task, dropping an object may be an essential skill.  Labeling and storing detailed sub-task episodes, even failures, can build a diverse skill set to be utilized later. The role of the data engine is to recover these hidden labels automatically or semi-automatically, turning physical behaviour into supervision.

This motivates the notion of embodied autolabelling. By embodied autolabelling, we mean the process of using physical sensing, temporal structure, and world knowledge to infer robot-relevant labels from behaviour without requiring manual annotation at every step. These labels may include when a task begins and ends, which object is being manipulated, where contact occurs, what state change is intended, whether progress is being made, and whether the outcome counts as success. Unlike ordinary video labelling, embodied autolabelling is not only semantic. It must be physically grounded: it should recover labels that are useful for control, such as grasp events, force-relevant contacts, object-centric state transitions, constraints, affordances, and failure modes.

Wearable sensing makes this problem especially interesting. A motion-capture or sensorimotor suit can provide structured signals that ordinary video lacks: body pose, hand trajectories, timing, contact events, tactile cues, force-related proxies, and possibly object interaction traces. This changes the role of human demonstrations. Instead of treating a human demonstration as only a video to imitate, we can treat it as a source of physically structured supervision. The human performs the task once, but the system extracts many labels: task phase boundaries, hand-object contacts, object state changes, intent, corrections, and candidate skill segments. These labels can then be used to train perception models, reward models, retargeting systems, world models, or robot policies. In this sense, embodied sensing is not merely a teleoperation interface; it is a labelling instrument for the physical world.

Human videos and data from wearable sensing suits also serve a dual purpose.  This data can teach robots about solving tasks, but it can also teach robots about humans themselves: how humans move, use their bodies and their environment, and interact with each other.  Robot intelligence should include a natural and collaborative working model of human behavior.  These sources of human data should be harnessed for training human-aware, human-compliant policies, and human collaborative policies.

The central challenge is that these labels are not independent. Task phase, contact, object state, action, and reward are coupled. A contact event matters because it changes an object state; an object state matters because it defines progress toward a task; a reward matters only relative to an inferred goal; and a robot action is useful only if it preserves the task-relevant structure of the human behaviour. A physical data engine must therefore integrate perception, tracking, temporal segmentation, contact inference, language grounding, and physical reasoning into a common representation. The goal is not to build a larger dataset in the ordinary sense, but to build a system that continuously transforms physical experience into reusable robot supervision.

\textbf{Raw Experiences are Heterogeneous and Asynchronous.} We start from a raw episode of physical experience. This episode may come from a robot rollout, human demonstration, wearable sensing system, internet video, simulation, or deployment trace. Crucially, the different streams may not be synchronised or sampled at the same frequency. We represent those by the following: 
\begin{equation*}
    \textbf{x} = \{(v_i, \tau_{i}^{(v)})_{i=1}^{T_v}, (m_j, \tau_j^{(m)})_{j=1}^{T_m}, (h_k, \tau_{k}^{h})_{k=1}^{T_h}, (r_{l}, \tau_{l}^{(r)})_{l=1}^{T_r}, \texttt{L}\}. 
\end{equation*}
Here, $(v_i, \tau_{i}^{(v)})$ denotes video frames with timestamps, $(m_j,\tau_{j}^{m})$ motion-capture, wearable, or body-pose measurements with timestamps, $(h_k,\tau_k^{(h)})$ tactile, force, contact, or hand-sensor readings with timestamps, $(r_l, \tau_l^{(r)})$ raw robot logs, if available, e.g., proprioception, deployment metadata, and $\texttt{L}$ language associated with the episode, such as an instruction, caption, task description, or human correction. 

Of course, not every episode contains all modalities. Internet video may contain only video and weak captions. A wearable-suit episode may contain video, pose, tactile, and language. A robot rollout may contain observations, actions, proprioception, and success/failure metadata. But, in its most general form, we say $\textbf{x} \in \mathcal{X}$, where $\mathcal{X}$ is the space of heterogeneous physical episodes. 

Because the streams are asynchronous, the first hidden object is an alignment between raw observations and a common physical timeline. Let $\zeta \in \{1, \dots, Z\}$ denote a latent event timeline. This does not have to be the same as the video rate or the robot control rate. It could be continuous time, a discretised timeline, or an event-based sequence. We introduce an alignment variable: $\mathcal{A}$, such that $\mathcal{A}:\{\tau_{i}^{(v)}, \tau_{j}^{(m)}, \tau_{k}^{(h)}, \tau_{l}^{(r)}\} \rightarrow \{1, \dots, Z\}.$
In words, $\mathcal{A}$ tells us which video frames, motion measurements, tactile events, robot logs, and language references correspond to the same underlying physical event. For example, it could be the case that: \texttt{video-frames:} 30-55, \texttt{motion-readings:} 102-180, and a \texttt{tactile-spike @:} 1.8 seconds, all map to the latent event of $\zeta =2:$ \texttt{contact-begins}.  In this sense, temporal alignment is not a preprocessing detail. It is, in fact, part of the embodied autolabelling problem. 

This immediately exposes the first interesting learning problem. Given a heterogeneous episode $\textbf{x}$, we want to infer a sequence of latent physical events $\zeta = 1, \dots, Z$, and for each event recover the variables that would make the episode useful for robot learning. We denote the latent structure associated with event $\zeta$ by: 
\begin{equation*}
    \textbf{z}_{\zeta} = [\textbf{s}_{\zeta}, \textbf{c}_{\zeta}, \boldsymbol{\phi}_{\zeta}, \textbf{u}_{\zeta}, 
\textbf{r}_{\zeta}],
\end{equation*}
where $\textbf{s}_{\zeta}$ is an object-centric physical state, $\textbf{c}_{\zeta}$ is a contact or interaction label, $\phi_{\zeta}$ is a task phase, $\textbf{u}_{\zeta}$ is a latent physical action or transition code, and $\textbf{r}_{\zeta}$ is a task-conditioned progress or reward signal. At the episode level, we also infer a goal $\textbf{g}$ and an outcome label $\textbf{y}$, such as success, failure, partial success, or unsafe execution. Thus, the full hidden explanation of an episode is: $\textbf{z}=[\textbf{z}_{1:Z}, \textbf{g}, \textbf{y}]$. The physical data engine can therefore be viewed as an inference model, $q_{\boldsymbol{\theta}}(\textbf{z}, \mathcal{A}|\textbf{x})$ which maps raw asynchronous multimodal experience into an aligned sequence of robot-relevant physical events. Importantly, $q_{\boldsymbol{\theta}}$ is not merely a perception model. It must jointly solve temporal alignment, event segmentation, object-state estimation, contact inference, phase recognition, latent-action discovery, reward grounding, and outcome prediction. In other words, the physical data engine tries to answer: what happened, when did it happen, which objects were involved, what physical change occurred, what task was being pursued, and whether that change constituted progress or failure.

To illustrate, consider a human wearing a sensing suit while placing a cup on a tray. The raw episode may contain video frames, body-pose measurements, hand trajectories, tactile spikes, and a language instruction. The inferred event sequence might be:
\begin{equation*}
    \zeta =1:  \texttt{reach-to-cup,} \ \ \zeta =2:  \texttt{contact-begins,} \ \ \zeta = 3: \texttt{grasp} \dots 
\end{equation*}
For each event, the engine should infer not only a semantic label, but a physical description: the cup pose, the hand-object contact, the relevant task phase, the latent transition being performed, and whether progress toward the goal is increasing. This is the distinction between ordinary video understanding and embodied autolabelling. A captioning model may say “a person places a cup on a tray”; a physical data engine should recover the sequence of physical events that could be retargeted, simulated, rewarded, or used to train a robot policy. 

A central open question is therefore how to learn the mapping: $\textbf{x} \rightarrow (\textbf{z}_{1:Z}, \textbf{g}, \textbf{y}, \mathcal{A})$, when most episodes provide only partial supervision. Some robot episodes contain actions but weak task labels. Some videos contain task semantics but no actions. Some wearable demonstrations contain motion and contact proxies but no robot embodiment. Some simulations provide full state but imperfect realism. A useful physical data engine must combine these sources by treating them as different views of the same underlying physical structure.

\subsection{Task-preserving Retargeting across Embodiments}
Inferring a structured event sequence from physical experience does not by itself produce a robot policy. A human demonstration, internet video, or wearable-sensor trace may reveal what happened physically — which object moved, where contact occurred, which task phase was executed, and whether progress was made — but it still does not specify how a particular robot should act. 

This is the embodiment gap. A human hand, parallel-jaw gripper, dexterous hand, mobile manipulator, quadruped, and humanoid all have different kinematics, dynamics, sensors, action spaces, contact surfaces, and failure modes. Therefore, the central question is not how to copy human motion, but how to preserve the task-relevant physical effect of that motion when executed by a different body. We call this task-preserving retargeting: the problem of mapping latent physical actions or human demonstrations into executable robot actions while preserving the intended effect on the world.

Let $\textbf{u}_{\zeta}$ denote the latent physical action inferred for event $\zeta$, and let $\textbf{s}_{\zeta}$ denote the corresponding object-centric state. For a robot embodiment e, retargeting seeks an executable action or skill: $\textbf{a}_{\zeta}^{(\text{embodied})} = f_{\psi}(\textbf{u}_\zeta, \textbf{s}_{\zeta}, \text{embodiment})$ such that the resulting robot-induced transition preserves the goal-relevant physical change $\Delta_{\textbf{g}}(\text{s}_{\zeta}, \textbf{a}_{\zeta}^{(\text{embodied})}) \approx \Delta_{\textbf{g}}(\textbf{s}_{\zeta}, \textbf{u}_{\zeta})$. Here, $\Delta_{\textbf{g}}$ denotes the task-relevant effect under goal $\textbf{g}$: drawer displacement for opening, object pose for placing, relative alignment for insertion, containment for packing, or contact state for grasping. This formulation makes clear why pose matching is insufficient. The correct retargeting target is not the human joint trajectory, but the physical transformation that matters for the task.

Retargeting can preserve different invariants. At the weakest level, it preserves pose: mapping human hand or arm motion to a robot end-effector trajectory. At a stronger level, it preserves contact: ensuring that the robot touches the relevant object surfaces at the relevant moments. Stronger still, it preserves object-state transitions: ensuring that the drawer opens, the cup is lifted, or the peg becomes aligned. The strongest form preserves intent or skill: the robot may use a different motion entirely, but accomplishes the same task under the same constraints. Generalist robotics will require retargeting to move up this hierarchy, from pose-preserving imitation to task-effect-preserving translation.

This view also clarifies why wearable sensing and embodied autolabelling are valuable. A suit or sensor-rich demonstration does not need to provide the final robot action directly. Instead, it can expose the intermediate variables required for task-preserving retargeting: hand-object contact, force-relevant events, object-state changes, task phase boundaries, and latent physical actions. These variables are more transferable than raw human joint angles and more informative than video captions. They form the bridge between human physical experience and robot-executable behaviour.

\subsection{Beyond Physics-Grounded World Models for Consequence Predictions}
Inferring physical events and retargeting them across embodiments still leaves one central problem: a robot must reason about consequences. A candidate action is useful only if the robot can anticipate what it will do to the world. Will the object move or slip? Will contact be established or lost? Will the drawer open or jam? Will the cup remain stable after release? Will the cloth deform in the intended direction? These are not merely visual questions. They require reasoning about geometry, contact, forces, constraints, material properties, and task progress. For this reason, the next generation of robot-learning systems requires physics-grounded world models: predictive models that estimate not only what the world may look like after an action, but what physically changes and why.

The role of a physics-grounded world model is therefore different from that of a generic video generator. A video model may produce plausible future frames, but a robot needs actionable predictions. It must know whether an action produces the desired object-state transition, whether a grasp is stable, whether a collision will occur, whether an insertion will fail because of misalignment, or whether an object will fall after being released. Thus, a robot world model should operate over structured physical variables whenever possible: object poses, spatial relations, contacts, constraints, velocities, forces, deformable states, and physical properties such as friction, mass, stiffness, or compliance. These variables are precisely the quantities that determine whether an imagined action can become a successful real-world behaviour.

We can write this role abstractly as consequence prediction. Given an object-centric state $\textbf{s}_{\zeta}$, a goal \textbf{g}, and either a latent physical action $\textbf{u}_{\zeta}$, or an embodiment-specific robot action $\textbf{a}_{\zeta}^{(\text{embodied})}$, the world model predicts the next physical state:
\begin{equation*}
    \textbf{s}_{\zeta+1} \sim p_{\omega}(\cdot| \textbf{s}_{\zeta}, \textbf{u}_{\zeta}, \textbf{g}).
\end{equation*}
For a specific robot embodiment, this would amount to: 
\begin{equation*}
    \textbf{s}_{\zeta+1} \sim p_{\omega}(\cdot| \textbf{s}_{\zeta}, \textbf{a}^{(\text{embodied})}_{\zeta}, \text{embodiement}, \textbf{g}).
\end{equation*}
The first form supports task-level reasoning: what physical transition should occur if the intended action is “pull”, “lift”, “insert”, or “place”? The second supports embodiment-specific planning: what will happen if this particular robot, with this morphology and controller, executes this action from this state? In both cases, the model should predict more than pixels. It should predict the physical variables that matter for control and reward.

This gives the world model a central role in the proposed stack. It can be used before action execution to evaluate candidate retargeted actions, during planning to search over alternatives, after failures to explain what went wrong, and during training to generate counterfactual experience. For example, if a human demonstration suggests the latent action “pull drawer outward”, a retargeting model may propose several robot motions. A physics-grounded world model can evaluate which motion is likely to establish the correct contact, apply force along the right direction, avoid collision, and produce the desired drawer displacement. Similarly, if a robot fails to insert a peg, the world model can help distinguish whether the failure came from poor alignment, insufficient force, unstable grasp, object geometry, or a wrong task phase.

The most important point is that consequence prediction should be task-conditioned. A world model does not need to predict every detail of the future equally well. It needs to predict the aspects of the future that are relevant to the task. For opening a drawer, drawer displacement and handle contact matter more than the exact background texture. For pouring, liquid state and container pose matter more than the appearance of the table. For folding cloth, deformable geometry and contact points matter more than pixel-perfect video reconstruction. This suggests that the objective of a robot world model should be aligned with downstream control, not only with visual reconstruction. The question is not “does the future look realistic?” but “does the prediction preserve the physical consequences that determine success or failure?”

This also clarifies why physics grounding matters. Purely learned predictors can exploit visual regularities without understanding the underlying constraints. They may generate futures in which objects interpenetrate, contacts occur without force, rigid objects deform unrealistically, or effects appear without plausible causes. Physics-grounded models reduce these failure modes by injecting structure: object permanence, geometric consistency, conservation laws, differentiable contact, learned or explicit material parameters, action-conditioned dynamics, and uncertainty estimates. The goal is not necessarily to build a perfect simulator. Rather, it is to learn a predictive model that is accurate where control needs accuracy and uncertain where the system lacks evidence.

In this view, world models are not separate from embodied autolabelling and retargeting. They complete the loop. The physical data engine infers what happened; retargeting proposes how a robot could reproduce the relevant physical effect; the world model predicts what would happen if the robot tried. These three components can improve one another. Autolabelled contact and object-state transitions provide supervision for the world model. The world model can detect inconsistent labels or impossible transitions. Retargeting can use world-model rollouts to choose actions that preserve task effects. Deployment failures can then be fed back into the data engine as new examples of what the model failed to predict.

The open challenge is to decide what representation such a world model should use. Pixel-space prediction offers broad coverage but weak physical abstraction. Object-centric models expose entities and relations but require reliable perception and tracking. 3D representations such as point clouds, meshes, neural fields, or Gaussian splats provide geometry but may still struggle with contact, force, and material response. Mechanics-based models encode physical laws but can be brittle when the environment is unknown or deformable. The most promising direction may be hybrid: learned models that combine 3D scene representations, object-centric structure, physics-inspired constraints, and data-driven residual dynamics. Such models would not merely imagine the future; they would provide physically meaningful counterfactuals for robot learning.

Physics-grounded world models therefore address a missing component in the path from physical experience to physical intelligence. They turn data into counterfactuals. Without them, a robot can only imitate what it has seen or execute what it has been commanded. With them, it can ask what would happen under alternative actions, bodies, contacts, and goals. But this promise depends on grounding: imagined futures are useful for robotics only when they preserve the physical structure that makes actions succeed or fail.
\subsection{Self-Improving Deployment Loops}
After a robot executes an action, the central question is no longer only what happened, but whether what happened was useful. A world model may predict that a cup will move, a drawer will open, or an object will fall; a retargeting system may produce a physically feasible action; and a policy may execute that action in the real world. But learning from the result requires a task-conditioned interpretation of the outcome. Did the action make progress? Did it solve the intended task? Did it fail because of perception, contact, force, timing, planning, or embodiment mismatch? Was the final state good or bad relative to the goal? These questions cannot be answered by a generic state evaluator. They require reward grounding: the ability to assign progress, success, and failure relative to the task being attempted. 

This is why robotic reward models should be task-conditioned. A physical state is not intrinsically successful or unsuccessful. The same state can mean different things depending on the goal: a cup resting on a table is success for “put the cup down”, failure for “pick up the cup”, and irrelevant for “open the drawer”. We can express this by writing reward as: $\textbf{r}_{\eta}(\textbf{s}_{\zeta}, \textbf{g}, \boldsymbol{\phi}_{\zeta})$, where $\textbf{s}_{\zeta}$ is the inferred physical state at event $\zeta$, $\textbf{g}$ is the task or goal, and $\boldsymbol{\phi}_{\zeta}$ is the task phase. In this view, reward is not merely a scalar attached to a state. It is an interpretation of physical progress under a goal. A good reward model should estimate whether the relevant contact occurred, whether the object moved in the intended way, whether the system entered a recoverable or unrecoverable failure mode, and whether the final configuration satisfies the task. 

This perspective also connects reward learning to video and language understanding. Many tasks can be recognised from the temporal structure of a demonstration or rollout: approaching, contacting, manipulating, releasing, verifying, recovering. Videos often reveal not only what task is being attempted, but also what progress and failure look like. Language can provide the task hypothesis; video and state estimates can provide evidence; human feedback can resolve ambiguity. Thus, reward grounding can be seen as task-conditioned physical interpretation: given a goal and an observed trajectory, infer which events count as progress, which count as failure, and which final states count as success. This is different from generic preference modelling. The reward must be tied to object states, contacts, constraints, and task phases.

This reward-grounding problem is what makes self-improving deployment possible. In a deployed robot system, every rollout should become more than a pass/fail record. It should become a labelled physical episode. A successful rollout provides examples of robust task completion. A failed rollout provides information about missing contact, wrong object state, unstable grasp, poor alignment, unsafe motion, or reward misinterpretation. A human correction provides a high-value supervision signal: it reveals not only that the robot was wrong, but often how the task should have proceeded. If these outcomes are fed back into the physical data engine, the system can update its reward model, retargeting model, world model, and policy.

The resulting loop is:  deploy policy $\rightarrow$
observe outcome $\rightarrow$ infer task-conditioned progress / success / failure $\rightarrow$
explain failure or correction $\rightarrow$
add grounded supervision to the data engine $\rightarrow$
update reward model, world model, retargeting, and policy $\rightarrow$
redeploy. This is the key difference between a robot that merely executes a trained policy and a robot-learning system that compounds over time. Without reward grounding, deployment traces are difficult to use: a failure is just a failed video, and a success is just an episode that happened to work. With task-conditioned reward grounding, deployment traces become structured supervision. The system can ask: which subgoal failed, which contact was missing, which object state was wrong, and what alternative action would have improved the outcome?

A self-improving deployment loop therefore requires three capabilities. First, it must monitor execution and detect meaningful events: contacts, state changes, subgoal completion, anomalies, and safety violations. Second, it must evaluate those events relative to the task, producing progress, reward, and failure labels. Third, it must route the resulting supervision to the right component: policy updates when the action was poor, world-model updates when the consequence prediction was wrong, retargeting updates when the physical effect was not preserved, and reward-model updates when success or failure was misclassified. This component-level credit assignment is essential. Otherwise, the system may know that a rollout failed but not what should change.

This closes the pipeline introduced in the paper. A physical data engine turns heterogeneous experience into latent physical events. Task-preserving retargeting maps those events to robot actions. Physics-grounded world models predict the consequences of those actions. Task-conditioned reward grounding interprets the outcomes. Deployment then supplies new episodes that re-enter the same pipeline. The long-term goal is a compounding physical-intelligence system: one in which every demonstration, video, simulation rollout, robot failure, and human correction becomes structured supervision for the next generation of robot behaviour.

\section{Conclusions}

This position paper has argued that the next stage of generalist robotics should not be framed only as a policy-scaling problem. Vision-language-action models, large robot datasets, simulation pipelines, and learned world models are all important parts of the emerging robotics stack. However, they do not by themselves solve the central bottleneck: how to convert the world’s broad, messy, and weakly labelled physical experience into supervision that a robot can actually use.

The key limitation is not simply a shortage of data. The world already contains enormous amounts of physical behaviour: humans manipulating objects, tools being used, factories operating, homes being organised, robots succeeding and failing, and simulations generating counterfactual interactions. The difficulty is that this experience does not arrive in the coordinate system of robot learning. It usually lacks robot actions, embodiment-specific constraints, task-phase labels, contact annotations, reward signals, and success or failure explanations. The missing problem is therefore grounding: transforming physical experience into robot-usable variables such as object states, contacts, latent actions, task phases, goals, rewards, and physically meaningful counterfactuals.

This perspective suggests that VLAs should be understood as one layer in a larger physical-intelligence stack. They provide a powerful policy interface between perception, language, and action, but their effectiveness depends on upstream and downstream mechanisms that make physical experience learnable. A robot-learning system must be able to autolabel heterogeneous behaviour, retarget task-relevant physical effects across embodiments, predict the consequences of candidate actions, and interpret deployment outcomes relative to the task being attempted. Without these components, larger policies may improve performance on curated robot-native datasets, but they will remain limited by the amount of experience that has already been manually or implicitly grounded for them.

This also suggests a different set of evaluation questions for generalist robotics. Instead of asking only whether a larger policy solves more tasks, we should ask whether a system can convert weaker sources of physical experience into useful supervision. Can it infer contacts, object-state changes, and task phases from human behaviour? Can it retarget a demonstrated physical effect to a new embodiment without merely copying pose? Can its world model predict the consequences that matter for success and failure, rather than only generating plausible future frames? Can its reward model distinguish progress, failure, recovery, and success relative to the current goal? Can deployment failures update the right component of the stack: the policy, the reward model, the world model, or the retargeting mechanism? These questions define the grounding agenda for robotics beyond VLA scaling.

The broader implication is that the next foundation model for robotics may not be a single monolithic model. It may instead be a compounding system: a physical data engine that turns heterogeneous experience into structured supervision; an embodiment interface that maps task-relevant effects to robot actions; a physics-grounded world model that generates actionable counterfactuals; and a task-conditioned deployment loop that converts successes, failures, and corrections into future improvement. In such a system, every human demonstration, internet video, simulation rollout, tactile trace, robot failure, and human correction becomes part of a growing supervision engine for physical intelligence.

Robots, therefore, need more than VLAs. They need the architectural pillars that make physical experience usable. Progress in robotics will depend not only on scaling policies but on building the grounding mechanisms that connect the world’s behavioural data to robot actions, rewards, models, and continual deployment. The central challenge for the field is to move from robot-native datasets to world-scale physical supervision, and from isolated policies to systems that learn from the physical world itself.
\bibliographystyle{plainnat}
\bibliography{refs}

@article{khazatsky2024droid,
  title={Droid: A large-scale in-the-wild robot manipulation dataset},
  author={Khazatsky, Alexander and Pertsch, Karl and Nair, Suraj and Balakrishna, Ashwin and Dasari, Sudeep and Karamcheti, Siddharth and Nasiriany, Soroush and Srirama, Mohan Kumar and Chen, Lawrence Yunliang and Ellis, Kirsty and others},
  journal={arXiv preprint arXiv:2403.12945},
  year={2024}
}

@article{feng2025spatial,
  title={Spatial-Aware VLA Pretraining through Visual-Physical Alignment from Human Videos},
  author={Feng, Yicheng and Zhang, Wanpeng and Wang, Ye and Luo, Hao and Yuan, Haoqi and Zheng, Sipeng and Lu, Zongqing},
  journal={arXiv preprint arXiv:2512.13080},
  year={2025}
}

@inproceedings{radosavovic2023real,
  title={Real-world robot learning with masked visual pre-training},
  author={Radosavovic, Ilija and Xiao, Tete and James, Stephen and Abbeel, Pieter and Malik, Jitendra and Darrell, Trevor},
  booktitle={Conference on Robot Learning},
  pages={416--426},
  year={2023},
  organization={PMLR}
}

@inproceedings{chen2025sarm,
  title={SARM: Stage-Aware Reward Modeling for Long Horizon Robot Manipulation},
  author={Chen, Qianzhong and Yu, Justin and Schwager, Mac and Abbeel, Pieter and Shentu, Yide and Wu, Philipp},
  booktitle = {{International Conference on Learning Representations (ICLR)}},
  journal={arXiv preprint arXiv:2509.25358},
  year={2026}
}

@article{low2025sous,
  title={Sous vide: Cooking visual drone navigation policies in a gaussian splatting vacuum},
  author={Low, JunEn and Adang, Maximilian and Yu, Javier and Nagami, Keiko and Schwager, Mac},
  journal={IEEE Robotics and Automation Letters},
  year={2025},
  publisher={IEEE}
}

@inproceedings{adang2025singer,
  title={SINGER: An onboard generalist vision-language navigation policy for drones},
  author={Adang, Maximilian and Low, JunEn and Shorinwa, Ola and Schwager, Mac},
  booktitle = {{In Proc. of the International Conference on Robotics and Automation (ICRA)}},
  journal={arXiv preprint arXiv:2509.18610},
  year={2026}
}

@inproceedings{chen2025grad,
  title={GRaD-Nav: Efficiently learning visual drone navigation with gaussian radiance fields and differentiable dynamics},
  author={Chen, Qianzhong and Sun, Jiankai and Gao, Naixiang and Low, JunEn and Chen, Timothy and Schwager, Mac},
  booktitle={2025 IEEE/RSJ International Conference on Intelligent Robots and Systems (IROS)},
  pages={7941--7948},
  year={2025},
  organization={IEEE}
}

@article{chen2025grad++,
  title={Grad-nav++: Vision-language model enabled visual drone navigation with gaussian radiance fields and differentiable dynamics},
  author={Chen, Qianzhong and Gao, Naixiang and Huang, Suning and Low, JunEn and Chen, Timothy and Sun, Jiankai and Schwager, Mac},
  journal={IEEE Robotics and Automation Letters},
  volume={11},
  number={2},
  pages={1418--1425},
  year={2025},
  publisher={IEEE}
}

@inproceedings{kumar2021rma,
  title={RMA: Rapid motor adaptation for legged robots},
  author={Kumar, Ashish and Fu, Zipeng and Pathak, Deepak and Malik, Jitendra},
  booktitle={Proceedings of Robotics: Science and Systems (RSS)},
  year={2021},
  address={Virtual Conference}
}

@inproceedings{wang2025phys2real,
  title={Phys2Real: Fusing VLM Priors with Interactive Online Adaptation for Uncertainty-Aware Sim-to-Real Manipulation},
  author={Wang, Maggie and Tian, Stephen and Swann, Aiden and Shorinwa, Ola and Wu, Jiajun and Schwager, Mac},
  booktitle = {{In Proc. of the International Conference on Robotics and Automation (ICRA)}},
  journal={arXiv preprint arXiv:2510.11689},
  year={2026}
}

@article{mei2025world,
  title={World Models That Know When They Don't Know: Controllable Video Generation with Calibrated Uncertainty},
  author={Mei, Zhiting and Yin, Tenny and Baker, Micah and Shorinwa, Ola and Majumdar, Anirudha},
  journal={arXiv preprint arXiv:2512.05927},
  year={2025}
}

@inproceedings{ward2026foundational,
  title={Foundational World Models Accurately Detect Bimanual Manipulator Failures},
  author={Ward, Isaac R and Ho, Michelle and Liu, Houjun and Feldman, Aaron and Vincent, Joseph and Kruse, Liam and Cheong, Sean and Eddy, Duncan and Kochenderfer, Mykel J and Schwager, Mac},
  booktitle = {{In Proc. of the International Conference on Robotics and Automation (ICRA)}},
  journal={arXiv preprint arXiv:2603.06987},
  year={2026}
}

@article{li2025uncertainty,
  title={Uncertainty-Aware Robotic World Model Makes Offline Model-Based Reinforcement Learning Work on Real Robots},
  author={Li, Chenhao and Krause, Andreas and Hutter, Marco},
  journal={arXiv preprint arXiv:2504.16680},
  year={2025}
}

@inproceedings{
ye2026world,
title={World Action Models are Zero-shot Policies},
author={Seonghyeon Ye and Yunhao Ge and Kaiyuan Zheng and Shenyuan Gao and Sihyun Yu and George Kurian and Suneel Indupuru and You Liang Tan and Chuning Zhu and Jiannan Xiang and Ayaan Naveed Malik and Kyungmin Lee and William Liang and Nadun Ranawaka Arachchige and Jiasheng Gu and Yinzhen Xu and Guanzhi Wang and Fengyuan Hu and Avnish Narayan and Johan Bjorck and Jing Wang and Gwanghyun Kim and Dantong Niu and Ruijie Zheng and Yuqi Xie and Jimmy Wu and Qi Wang and Danfei Xu and Yilun Du and Ryan Julian and Yevgen Chebotar and Scott Reed and Jan Kautz and Yuke Zhu and Linxi Fan and Joel Jang},
booktitle={ICLR 2026 the 2nd Workshop on World Models: Understanding, Modelling and Scaling},
year={2026},
url={https://openreview.net/forum?id=cd33uUB609}
}

@inproceedings{li2025unified,
  title={Unified video action model},
  author={Li, Shuang and Gao, Yihuai and Sadigh, Dorsa and Song, Shuran},
  booktitle = {{Robotics: Science and Systems (RSS)}},
  journal={arXiv preprint arXiv:2503.00200},
  year={2026}
}

@article{fang2023rh20t,
  title={Rh20t: A comprehensive robotic dataset for learning diverse skills in one-shot},
  author={Fang, Hao-Shu and Fang, Hongjie and Tang, Zhenyu and Liu, Jirong and Wang, Chenxi and Wang, Junbo and Zhu, Haoyi and Lu, Cewu},
  journal={arXiv preprint arXiv:2307.00595},
  year={2023}
}

@misc{figure2025helix,
  author       = {{Figure AI}},
  title        = {{Helix}: A Vision-Language-Action Model for Generalist Humanoid Control},
  year         = {2025},
  month        = feb,
  day          = {20},
  howpublished = {\url{https://www.figure.ai/news/helix}},
  note         = {Accessed: 2026-04-28}
}

@article{bai2026hex,
  title={HEX: Humanoid-Aligned Experts for Cross-Embodiment Whole-Body Manipulation},
  author={Bai, Shuanghao and Li, Meng and Lv, Xinyuan and Wang, Jiawei and Wang, Xinhua and Liao, Fei and Hou, Chengkai and Gu, Langzhe and Zhou, Wanqi and Wu, Kun and others},
  journal={arXiv preprint arXiv:2604.07993},
  year={2026}
}

@misc{nai2026humanoidmanipulationinterfacehumanoid,
      title={Humanoid Manipulation Interface: Humanoid Whole-Body Manipulation from Robot-Free Demonstrations}, 
      author={Ruiqian Nai and Boyuan Zheng and Junming Zhao and Haodong Zhu and Sicong Dai and Zunhao Chen and Yihang Hu and Yingdong Hu and Tong Zhang and Chuan Wen and Yang Gao},
      year={2026},
      eprint={2602.06643},
      archivePrefix={arXiv},
      primaryClass={cs.RO},
      url={https://arxiv.org/abs/2602.06643}, 
}

@misc{skild2025brain,
  author       = {{Skild AI}},
  title        = {Building the General-Purpose Robotic Brain},
  year         = {2025},
  month        = jul,
  day          = {29},
  howpublished = {\url{https://www.skild.ai/blogs/building-the-general-purpose-robotic-brain}},
  note         = {Accessed: 2026-04-28}
}

@article{jiang2025wholebodyvla,
  title={Wholebodyvla: Towards unified latent vla for whole-body loco-manipulation control},
  author={Jiang, Haoran and Chen, Jin and Bu, Qingwen and Chen, Li and Shi, Modi and Zhang, Yanjie and Li, Delong and Suo, Chuanzhe and Wang, Chuang and Peng, Zhihui and others},
  journal={arXiv preprint arXiv:2512.11047},
  year={2025}
}

@article{xue2025leverb,
  title={Leverb: Humanoid whole-body control with latent vision-language instruction},
  author={Xue, Haoru and Huang, Xiaoyu and Niu, Dantong and Liao, Qiayuan and Kragerud, Thomas and Gravdahl, Jan Tommy and Peng, Xue Bin and Shi, Guanya and Darrell, Trevor and Sreenath, Koushil and others},
  journal={arXiv preprint arXiv:2506.13751},
  year={2025}
}

@article{song2025avi,
  title={Avi: Action from Volumetric Inference},
  author={Song, Harris and Le, Long},
  journal={arXiv preprint arXiv:2510.21746},
  year={2025}
}

@inproceedings{huang2026graphcot,
  title={Graphcot-vla: A 3d spatial-aware reasoning vision-language-action model for robotic manipulation with ambiguous instructions},
  author={Huang, Helong and Cen, Min and Tan, Kai and Quan, Xingyue and Huang, Guowei and Zhang, Hong},
  booktitle={Proceedings of the AAAI Conference on Artificial Intelligence},
  volume={40},
  number={22},
  pages={18324--18332},
  year={2026}
}

@article{sun2025geovla,
  title={Geovla: Empowering 3d representations in vision-language-action models},
  author={Sun, Lin and Xie, Bin and Liu, Yingfei and Shi, Hao and Wang, Tiancai and Cao, Jiale},
  journal={arXiv preprint arXiv:2508.09071},
  year={2025}
}

@inproceedings{li20253ds,
  title={3ds-vla: A 3d spatial-aware vision language action model for robust multi-task manipulation},
  author={Li, Xiaoqi and Heng, Liang and Liu, Jiaming and Shen, Yan and Gu, Chenyang and Liu, Zhuoyang and Chen, Hao and Han, Nuowei and Zhang, Renrui and Tang, Hao and others},
  booktitle={9th Annual Conference on Robot Learning},
  year={2025}
}

@misc{qu2025spatialvlaexploringspatialrepresentations,
      title={SpatialVLA: Exploring Spatial Representations for Visual-Language-Action Model}, 
      author={Delin Qu and Haoming Song and Qizhi Chen and Yuanqi Yao and Xinyi Ye and Yan Ding and Zhigang Wang and JiaYuan Gu and Bin Zhao and Dong Wang and Xuelong Li},
      year={2025},
      eprint={2501.15830},
      archivePrefix={arXiv},
      primaryClass={cs.RO},
      url={https://arxiv.org/abs/2501.15830}, 
}

@article{moodley2024multi,
  title={Multi-State-Action Tokenisation in Decision Transformers for Multi-Discrete Action Spaces},
  author={Moodley, Perusha and Kaushik, Pramod and Thambi, Dhillu and Trovinger, Mark and Paruchuri, Praveen and Hong, Xia and Rosman, Benjamin},
  journal={arXiv preprint arXiv:2407.01310},
  year={2024}
}

@article{zhong2025survey,
  title={A survey on vision-language-action models: An action tokenization perspective},
  author={Zhong, Yifan and Bai, Fengshuo and Cai, Shaofei and Huang, Xuchuan and Chen, Zhang and Zhang, Xiaowei and Wang, Yuanfei and Guo, Shaoyang and Guan, Tianrui and Lui, Ka Nam and others},
  journal={arXiv preprint arXiv:2507.01925},
  year={2025}
}

@article{dong2026actioncodec,
  title={ActionCodec: What Makes for Good Action Tokenizers},
  author={Dong, Zibin and Liu, Yicheng and Zhang, Shiduo and Ye, Baijun and Yuan, Yifu and Ni, Fei and Gong, Jingjing and Qiu, Xipeng and Zhao, Hang and Li, Yinchuan and others},
  journal={arXiv preprint arXiv:2602.15397},
  year={2026}
}

@article{liu2025faster,
  title={FASTer: Toward Efficient Autoregressive Vision Language Action Modeling via Neural Action Tokenization},
  author={Liu, Yicheng and Zhang, Shiduo and Dong, Zibin and Ye, Baijun and Yuan, Tianyuan and Yu, Xiaopeng and Yin, Linqi and Lu, Chenhao and Shi, Junhao and Yu, Luca Jiang-Tao and others},
  journal={arXiv preprint arXiv:2512.04952},
  year={2025}
}

@article{pertsch2025fast,
  title={Fast: Efficient action tokenization for vision-language-action models},
  author={Pertsch, Karl and Stachowicz, Kyle and Ichter, Brian and Driess, Danny and Nair, Suraj and Vuong, Quan and Mees, Oier and Finn, Chelsea and Levine, Sergey},
  journal={arXiv preprint arXiv:2501.09747},
  year={2025}
}

@article{liu2024robomamba,
  title={Robomamba: Efficient vision-language-action model for robotic reasoning and manipulation},
  author={Liu, Jiaming and Liu, Mengzhen and Wang, Zhenyu and An, Pengju and Li, Xiaoqi and Zhou, Kaichen and Yang, Senqiao and Zhang, Renrui and Guo, Yandong and Zhang, Shanghang},
  journal={Advances in Neural Information Processing Systems},
  volume={37},
  pages={40085--40110},
  year={2024}
}

@article{li2024cogact,
  title={Cogact: A foundational vision-language-action model for synergizing cognition and action in robotic manipulation},
  author={Li, Qixiu and Liang, Yaobo and Wang, Zeyu and Luo, Lin and Chen, Xi and Liao, Mozheng and Wei, Fangyun and Deng, Yu and Xu, Sicheng and Zhang, Yizhong and others},
  journal={arXiv preprint arXiv:2411.19650},
  year={2024}
}

@article{liu2024rdt,
  title={Rdt-1b: a diffusion foundation model for bimanual manipulation},
  author={Liu, Songming and Wu, Lingxuan and Li, Bangguo and Tan, Hengkai and Chen, Huayu and Wang, Zhengyi and Xu, Ke and Su, Hang and Zhu, Jun},
  journal={arXiv preprint arXiv:2410.07864},
  year={2024}
}

@article{zhang2024vla,
  title={Vla-3d: A dataset for 3d semantic scene understanding and navigation},
  author={Zhang, Haochen and Zantout, Nader and Kachana, Pujith and Wu, Zongyuan and Zhang, Ji and Wang, Wenshan},
  journal={arXiv preprint arXiv:2411.03540},
  year={2024}
}

@article{patratskiy2025spatial,
  title={Spatial traces: Enhancing vla models with spatial-temporal understanding},
  author={Patratskiy, Maxim A and Kovalev, Alexey K and Panov, Aleksandr I},
  journal={Optical Memory and Neural Networks},
  volume={34},
  number={Suppl 1},
  pages={S72--S82},
  year={2025},
  publisher={Springer}
}

@article{qu2025spatialvla,
  title={Spatialvla: Exploring spatial representations for visual-language-action model},
  author={Qu, Delin and Song, Haoming and Chen, Qizhi and Yao, Yuanqi and Ye, Xinyi and Ding, Yan and Wang, Zhigang and Gu, JiaYuan and Zhao, Bin and Wang, Dong and others},
  journal={arXiv preprint arXiv:2501.15830},
  year={2025}
}

@article{reed2022generalist,
  title={A generalist agent},
  author={Reed, Scott and Zolna, Konrad and Parisotto, Emilio and Colmenarejo, Sergio Gomez and Novikov, Alexander and Barth-Maron, Gabriel and Gimenez, Mai and Sulsky, Yury and Kay, Jackie and Springenberg, Jost Tobias and others},
  journal={arXiv preprint arXiv:2205.06175},
  year={2022}
}

@article{team2025gemini,
  title={Gemini robotics: Bringing ai into the physical world},
  author={Team, Gemini Robotics and Abeyruwan, Saminda and Ainslie, Joshua and Alayrac, Jean-Baptiste and Arenas, Montserrat Gonzalez and Armstrong, Travis and Balakrishna, Ashwin and Baruch, Robert and Bauza, Maria and Blokzijl, Michiel and others},
  journal={arXiv preprint arXiv:2503.20020},
  year={2025}
}

@article{pinwm2025,
  title={PIN-WM: Learning Physics-INformed World Models for Non-Prehensile Manipulation},
  author={Authors of PIN-WM},
  journal={arXiv preprint arXiv:2504.16693},
  year={2025}
}

@misc{wang2026contactgaussianwmlearningphysicsgroundedworld,
      title={ContactGaussian-WM: Learning Physics-Grounded World Model from Videos}, 
      author={Meizhong Wang and Wanxin Jin and Kun Cao and Lihua Xie and Yiguang Hong},
      year={2026},
      eprint={2602.11021},
      archivePrefix={arXiv},
      primaryClass={cs.RO},
      url={https://arxiv.org/abs/2602.11021}, 
}

@article{bjorck2025gr00t,
  title={Gr00t n1: An open foundation model for generalist humanoid robots},
  author={Bjorck, Johan and Casta{\~n}eda, Fernando and Cherniadev, Nikita and Da, Xingye and Ding, Runyu and Fan, Linxi and Fang, Yu and Fox, Dieter and Hu, Fengyuan and Huang, Spencer and others},
  journal={arXiv preprint arXiv:2503.14734},
  year={2025}
}

@article{lu2025gwm,
  title={GWM: Towards Scalable Gaussian World Models for Robotic Manipulation},
  author={Lu, Guanxing and Jia, Baoxiong and Li, Puhao and Chen, Yixin and Wang, Ziwei and Tang, Yansong and Huang, Siyuan},
  journal={arXiv preprint arXiv:2508.17600},
  year={2025}
}

@inproceedings{polymenakos2019safe,
  title={Safe policy search using Gaussian process models},
  author={Polymenakos, Kyriakos and Abate, Alessandro and Roberts, Stephen},
  booktitle={Proceedings of the 18th international conference on autonomous agents and multiagent systems},
  pages={1565--1573},
  year={2019}
}

@misc{xie2024physgaussianphysicsintegrated3dgaussians,
      title={PhysGaussian: Physics-Integrated 3D Gaussians for Generative Dynamics}, 
      author={Tianyi Xie and Zeshun Zong and Yuxing Qiu and Xuan Li and Yutao Feng and Yin Yang and Chenfanfu Jiang},
      year={2024},
      eprint={2311.12198},
      archivePrefix={arXiv},
      primaryClass={cs.GR},
      url={https://arxiv.org/abs/2311.12198}, 
}

@misc{sanchezgonzalez2020learningsimulatecomplexphysics,
      title={Learning to Simulate Complex Physics with Graph Networks}, 
      author={Alvaro Sanchez-Gonzalez and Jonathan Godwin and Tobias Pfaff and Rex Ying and Jure Leskovec and Peter W. Battaglia},
      year={2020},
      eprint={2002.09405},
      archivePrefix={arXiv},
      primaryClass={cs.LG},
      url={https://arxiv.org/abs/2002.09405}, 
}

@misc{sanchezgonzalez2018graphnetworkslearnablephysics,
      title={Graph networks as learnable physics engines for inference and control}, 
      author={Alvaro Sanchez-Gonzalez and Nicolas Heess and Jost Tobias Springenberg and Josh Merel and Martin Riedmiller and Raia Hadsell and Peter Battaglia},
      year={2018},
      eprint={1806.01242},
      archivePrefix={arXiv},
      primaryClass={cs.LG},
      url={https://arxiv.org/abs/1806.01242}, 
}

@misc{chang2017compositionalobjectbasedapproachlearning,
      title={A Compositional Object-Based Approach to Learning Physical Dynamics}, 
      author={Michael B. Chang and Tomer Ullman and Antonio Torralba and Joshua B. Tenenbaum},
      year={2017},
      eprint={1612.00341},
      archivePrefix={arXiv},
      primaryClass={cs.AI},
      url={https://arxiv.org/abs/1612.00341}, 
}

@article{Inter,
  author       = {Peter W. Battaglia and
                  Razvan Pascanu and
                  Matthew Lai and
                  Danilo Jimenez Rezende and
                  Koray Kavukcuoglu},
  title        = {Interaction Networks for Learning about Objects, Relations and Physics},
  journal      = {CoRR},
  volume       = {abs/1612.00222},
  year         = {2016},
  url          = {http://arxiv.org/abs/1612.00222},
  eprinttype   = {arXiv},
  eprint       = {1612.00222},
  bibsource    = {dblp computer science bibliography, https://dblp.org}
}

@article{SODEs,
  author       = {Yaofeng Desmond Zhong and
                  Biswadip Dey and
                  Amit Chakraborty},
  title        = {Symplectic ODE-Net: Learning Hamiltonian Dynamics with Control},
  journal      = {CoRR},
  volume       = {abs/1909.12077},
  year         = {2019},
  url          = {http://arxiv.org/abs/1909.12077},
  eprinttype   = {arXiv},
  eprint       = {1909.12077},
  bibsource    = {dblp computer science bibliography, https://dblp.org}
}

@article{LNNs,
  author       = {Miles D. Cranmer and
                  Sam Greydanus and
                  Stephan Hoyer and
                  Peter W. Battaglia and
                  David N. Spergel and
                  Shirley Ho},
  title        = {Lagrangian Neural Networks},
  journal      = {CoRR},
  volume       = {abs/2003.04630},
  year         = {2020},
  url          = {https://arxiv.org/abs/2003.04630},
  eprinttype   = {arXiv},
  eprint       = {2003.04630},
  bibsource    = {dblp computer science bibliography, https://dblp.org}
}

@article{HNNs,
  author       = {Sam Greydanus and
                  Misko Dzamba and
                  Jason Yosinski},
  title        = {Hamiltonian Neural Networks},
  journal      = {CoRR},
  volume       = {abs/1906.01563},
  year         = {2019},
  url          = {http://arxiv.org/abs/1906.01563},
  eprinttype   = {arXiv},
  eprint       = {1906.01563},
  bibsource    = {dblp computer science bibliography, https://dblp.org}
}

@article{janDeepLagrange,
  author       = {Michael Lutter and
                  Christian Ritter and
                  Jan Peters},
  title        = {Deep Lagrangian Networks: Using Physics as Model Prior for Deep Learning},
  journal      = {CoRR},
  volume       = {abs/1907.04490},
  year         = {2019},
  url          = {http://arxiv.org/abs/1907.04490},
  eprinttype   = {arXiv},
  eprint       = {1907.04490},
  bibsource    = {dblp computer science bibliography, https://dblp.org}
}

@misc{assran2025vjepa2selfsupervisedvideo,
      title={V-JEPA 2: Self-Supervised Video Models Enable Understanding, Prediction and Planning}, 
      author={Mido Assran and Adrien Bardes and David Fan and Quentin Garrido and Russell Howes and Mojtaba and Komeili and Matthew Muckley and Ammar Rizvi and Claire Roberts and Koustuv Sinha and Artem Zholus and Sergio Arnaud and Abha Gejji and Ada Martin and Francois Robert Hogan and Daniel Dugas and Piotr Bojanowski and Vasil Khalidov and Patrick Labatut and Francisco Massa and Marc Szafraniec and Kapil Krishnakumar and Yong Li and Xiaodong Ma and Sarath Chandar and Franziska Meier and Yann LeCun and Michael Rabbat and Nicolas Ballas},
      year={2025},
      eprint={2506.09985},
      archivePrefix={arXiv},
      primaryClass={cs.AI},
      url={https://arxiv.org/abs/2506.09985}, 
}

@inproceedings{assran2023self,
  title={Self-supervised learning from images with a joint-embedding predictive architecture},
  author={Assran, Mahmoud and Duval, Quentin and Misra, Ishan and Bojanowski, Piotr and Vincent, Pascal and Rabbat, Michael and LeCun, Yann and Ballas, Nicolas},
  booktitle={Proceedings of the IEEE/CVF conference on computer vision and pattern recognition},
  pages={15619--15629},
  year={2023}
}

@article{lecun2022path,
  title={A path towards autonomous machine intelligence version 0.9. 2, 2022-06-27},
  author={LeCun, Yann and others},
  journal={Open Review},
  volume={62},
  number={1},
  pages={1--62},
  year={2022}
}

@article{cowen2022samba,
  title={Samba: Safe model-based \& active reinforcement learning},
  author={Cowen-Rivers, Alexander I and Palenicek, Daniel and Moens, Vincent and Abdullah, Mohammed Amin and Sootla, Aivar and Wang, Jun and Bou-Ammar, Haitham},
  journal={Machine Learning},
  volume={111},
  number={1},
  pages={173--203},
  year={2022},
  publisher={Springer}
}

@inproceedings{deisenroth2011pilco,
  title={PILCO: A model-based and data-efficient approach to policy search},
  author={Deisenroth, Marc and Rasmussen, Carl E},
  booktitle={Proceedings of the 28th International Conference on machine learning (ICML-11)},
  pages={465--472},
  year={2011}
}

@article{Kober2013IJRR,
  author = {Kober, Jens and Bagnell, J. Andrew and Peters, Jan},
  title = {Reinforcement Learning in Robotics: A Survey},
  journal = {International Journal of Robotics Research},
  volume = {32},
  number = {11},
  pages = {1238--1274},
  year = {2013},
  doi = {10.1177/0278364913495721},
  url = {http://sagepub.com}
}

@misc{huang2025particleformer3dpointcloud,
      title={ParticleFormer: A 3D Point Cloud World Model for Multi-Object, Multi-Material Robotic Manipulation}, 
      author={Suning Huang and Qianzhong Chen and Xiaohan Zhang and Jiankai Sun and Mac Schwager},
      year={2025},
      eprint={2506.23126},
      archivePrefix={arXiv},
      primaryClass={cs.RO},
      url={https://arxiv.org/abs/2506.23126}, 
}

@misc{huang2026pointworldscaling3dworld,
      title={PointWorld: Scaling 3D World Models for In-The-Wild Robotic Manipulation}, 
      author={Wenlong Huang and Yu-Wei Chao and Arsalan Mousavian and Ming-Yu Liu and Dieter Fox and Kaichun Mo and Li Fei-Fei},
      year={2026},
      eprint={2601.03782},
      archivePrefix={arXiv},
      primaryClass={cs.RO},
      url={https://arxiv.org/abs/2601.03782}, 
}

@misc{jeong2025objectcentricworldmodellanguageguided,
      title={Object-Centric World Model for Language-Guided Manipulation}, 
      author={Youngjoon Jeong and Junha Chun and Soonwoo Cha and Taesup Kim},
      year={2025},
      eprint={2503.06170},
      archivePrefix={arXiv},
      primaryClass={cs.AI},
      url={https://arxiv.org/abs/2503.06170}, 
}

@misc{ferraro2023focusobjectcentricworldmodels,
      title={FOCUS: Object-Centric World Models for Robotics Manipulation}, 
      author={Stefano Ferraro and Pietro Mazzaglia and Tim Verbelen and Bart Dhoedt},
      year={2023},
      eprint={2307.02427},
      archivePrefix={arXiv},
      primaryClass={cs.RO},
      url={https://arxiv.org/abs/2307.02427}, 
}

@inproceedings{bruce2024genie,
  title={Genie: Generative Interactive Environments},
  author={Bruce, Jake and Dennis, Michael and Edwards, Ashley and Parker-Holder, Jack and Shi, Yuge and Hughes, Edward and Lai, Matthew and Mavalankar, Aditi and Steigerwald, Richie and Apps, Chris and others},
  booktitle={Forty-first International Conference on Machine Learning},
  year={2024},
  url={https://openreview.net/forum?id=bJbSbJskOS}
}

@misc{yang2024learninginteractiverealworldsimulators,
      title={Learning Interactive Real-World Simulators}, 
      author={Sherry Yang and Yilun Du and Kamyar Ghasemipour and Jonathan Tompson and Leslie Kaelbling and Dale Schuurmans and Pieter Abbeel},
      year={2024},
      eprint={2310.06114},
      archivePrefix={arXiv},
      primaryClass={cs.AI},
      url={https://arxiv.org/abs/2310.06114}, 
}

@misc{zhou2024robodreamerlearningcompositionalworld,
      title={RoboDreamer: Learning Compositional World Models for Robot Imagination}, 
      author={Siyuan Zhou and Yilun Du and Jiaben Chen and Yandong Li and Dit-Yan Yeung and Chuang Gan},
      year={2024},
      eprint={2404.12377},
      archivePrefix={arXiv},
      primaryClass={cs.RO},
      url={https://arxiv.org/abs/2404.12377}, 
}

@misc{wu2022daydreamerworldmodelsphysical,
      title={DayDreamer: World Models for Physical Robot Learning}, 
      author={Philipp Wu and Alejandro Escontrela and Danijar Hafner and Ken Goldberg and Pieter Abbeel},
      year={2022},
      eprint={2206.14176},
      archivePrefix={arXiv},
      primaryClass={cs.RO},
      url={https://arxiv.org/abs/2206.14176}, 
}

@article{hafner2025mastering,
  title={Mastering diverse control tasks through world models},
  author={Hafner, Danijar and Pasukonis, Jurgis and Ba, Jimmy and Lillicrap, Timothy},
  journal={Nature},
  volume={640},
  number={8059},
  pages={647--653},
  year={2025},
  publisher={Nature Publishing Group UK London}
}

@article{liu2025timerewarder,
  title={TimeRewarder: Learning Dense Reward from Passive Videos via Frame-wise Temporal Distance},
  author={Liu, Yuyang and Wen, Chuan and Hu, Yihang and Jayaraman, Dinesh and Gao, Yang},
  journal={arXiv preprint arXiv:2509.26627},
  year={2025}
}

@misc{james2019rlbenchrobotlearningbenchmark,
      title={RLBench: The Robot Learning Benchmark \& Learning Environment}, 
      author={Stephen James and Zicong Ma and David Rovick Arrojo and Andrew J. Davison},
      year={2019},
      eprint={1909.12271},
      archivePrefix={arXiv},
      primaryClass={cs.RO},
      url={https://arxiv.org/abs/1909.12271}, 
}

@article{DreamerV2,
author= {Danijar Hafner and
                  Timothy P. Lillicrap and
                  Mohammad Norouzi and
                  Jimmy Ba},
  title        = {Mastering Atari with Discrete World Models},
  journal      = {CoRR},
  volume       = {abs/2010.02193},
  year         = {2020},
  url          = {https://arxiv.org/abs/2010.02193},
  eprinttype   = {arXiv},
  eprint       = {2010.02193},
  bibsource    = {dblp computer science bibliography, https://dblp.org}
}

@inproceedings{hafner2019planet,
  title={Learning Latent Dynamics for Planning from Pixels},
  author={Hafner, Danijar and Lillicrap, Timothy and Fischer, Ian and 
          Villegas, Ruben and Ha, David and Lee, Honglak and Davidson, James},
  booktitle={International Conference on Machine Learning (ICML)},
  year={2019},
  pages={2555--2565},
  url={https://arxiv.org/abs/1811.04551}
}

@misc{ha2018recurrentworldmodelsfacilitate,
      title={Recurrent World Models Facilitate Policy Evolution}, 
      author={David Ha and Jürgen Schmidhuber},
      year={2018},
      eprint={1809.01999},
      archivePrefix={arXiv},
      primaryClass={cs.LG},
      url={https://arxiv.org/abs/1809.01999}, 
}

@article{Dreamerv1,
  author       = {Danijar Hafner and
                  Timothy P. Lillicrap and
                  Jimmy Ba and
                  Mohammad Norouzi},
  title        = {Dream to Control: Learning Behaviors by Latent Imagination},
  journal      = {CoRR},
  volume       = {abs/1912.01603},
  year         = {2019},
  url          = {http://arxiv.org/abs/1912.01603},
  eprinttype   = {arXiv},
  eprint       = {1912.01603},
  bibsource    = {dblp computer science bibliography, https://dblp.org}
}

@article{Jurg,
  doi = {10.5281/ZENODO.1207631},
  url = {https://zenodo.org/record/1207631},
  author = {Ha, David and Schmidhuber, Jürgen},
  title = {World Models},
  publisher = {Zenodo},
  year = {2018},
  copyright = {Creative Commons Attribution 4.0}
}

@misc{rudin2022learningwalkminutesusing,
      title={Learning to Walk in Minutes Using Massively Parallel Deep Reinforcement Learning}, 
      author={Nikita Rudin and David Hoeller and Philipp Reist and Marco Hutter},
      year={2022},
      eprint={2109.11978},
      archivePrefix={arXiv},
      primaryClass={cs.RO},
      url={https://arxiv.org/abs/2109.11978}, 
}

@article{Hwangbo_2019,
   title={Learning agile and dynamic motor skills for legged robots},
   volume={4},
   ISSN={2470-9476},
   url={http://dx.doi.org/10.1126/scirobotics.aau5872},
   DOI={10.1126/scirobotics.aau5872},
   number={26},
   journal={Science Robotics},
   publisher={American Association for the Advancement of Science (AAAS)},
   author={Hwangbo, Jemin and Lee, Joonho and Dosovitskiy, Alexey and Bellicoso, Dario and Tsounis, Vassilios and Koltun, Vladlen and Hutter, Marco},
   year={2019},
   month=Jan }

@misc{li2025robogsimreal2sim2realroboticgaussian,
      title={RoboGSim: A Real2Sim2Real Robotic Gaussian Splatting Simulator}, 
      author={Xinhai Li and Jialin Li and Ziheng Zhang and Rui Zhang and Fan Jia and Tiancai Wang and Haoqiang Fan and Kuo-Kun Tseng and Ruiping Wang},
      year={2025},
      eprint={2411.11839},
      archivePrefix={arXiv},
      primaryClass={cs.RO},
      url={https://arxiv.org/abs/2411.11839}, 
}

@misc{muratore2022robotlearningrandomizedsimulations,
      title={Robot Learning from Randomized Simulations: A Review}, 
      author={Fabio Muratore and Fabio Ramos and Greg Turk and Wenhao Yu and Michael Gienger and Jan Peters},
      year={2022},
      eprint={2111.00956},
      archivePrefix={arXiv},
      primaryClass={cs.RO},
      url={https://arxiv.org/abs/2111.00956}, 
}

@misc{zhang2025realtosimrobotpolicyevaluation,
      title={Real-to-Sim Robot Policy Evaluation with Gaussian Splatting Simulation of Soft-Body Interactions}, 
      author={Kaifeng Zhang and Shuo Sha and Hanxiao Jiang and Matthew Loper and Hyunjong Song and Guangyan Cai and Zhuo Xu and Xiaochen Hu and Changxi Zheng and Yunzhu Li},
      year={2025},
      eprint={2511.04665},
      archivePrefix={arXiv},
      primaryClass={cs.RO},
      url={https://arxiv.org/abs/2511.04665}, 
}

@misc{abouchakra2025realissimbridgingsimtorealgap,
      title={Real-is-Sim: Bridging the Sim-to-Real Gap with a Dynamic Digital Twin}, 
      author={Jad Abou-Chakra and Lingfeng Sun and Krishan Rana and Brandon May and Karl Schmeckpeper and Niko Suenderhauf and Maria Vittoria Minniti and Laura Herlant},
      year={2025},
      eprint={2504.03597},
      archivePrefix={arXiv},
      primaryClass={cs.RO},
      url={https://arxiv.org/abs/2504.03597}, 
}

@misc{torne2024reconcilingrealitysimulationrealtosimtoreal,
      title={Reconciling Reality through Simulation: A Real-to-Sim-to-Real Approach for Robust Manipulation}, 
      author={Marcel Torne and Anthony Simeonov and Zechu Li and April Chan and Tao Chen and Abhishek Gupta and Pulkit Agrawal},
      year={2024},
      eprint={2403.03949},
      archivePrefix={arXiv},
      primaryClass={cs.RO},
      url={https://arxiv.org/abs/2403.03949}, 
}

@misc{wu2025rlgsbridge3dgaussiansplatting,
      title={RL-GSBridge: 3D Gaussian Splatting Based Real2Sim2Real Method for Robotic Manipulation Learning}, 
      author={Yuxuan Wu and Lei Pan and Wenhua Wu and Guangming Wang and Yanzi Miao and Fan Xu and Hesheng Wang},
      year={2025},
      eprint={2409.20291},
      archivePrefix={arXiv},
      primaryClass={cs.RO},
      url={https://arxiv.org/abs/2409.20291}, 
}

@misc{wang2024robogenunleashinginfinitedata,
      title={RoboGen: Towards Unleashing Infinite Data for Automated Robot Learning via Generative Simulation}, 
      author={Yufei Wang and Zhou Xian and Feng Chen and Tsun-Hsuan Wang and Yian Wang and Katerina Fragkiadaki and Zackory Erickson and David Held and Chuang Gan},
      year={2024},
      eprint={2311.01455},
      archivePrefix={arXiv},
      primaryClass={cs.RO},
      url={https://arxiv.org/abs/2311.01455}, 
}

@article{nasiriany2026robocasa365,
  title={Robocasa365: A large-scale simulation framework for training and benchmarking generalist robots},
  author={Nasiriany, Soroush and Nasiriany, Sepehr and Maddukuri, Abhiram and Zhu, Yuke},
  journal={arXiv preprint arXiv:2603.04356},
  year={2026}
}

@article{nasiriany2024robocasa,
  title={Robocasa: Large-scale simulation of everyday tasks for generalist robots},
  author={Nasiriany, Soroush and Maddukuri, Abhiram and Zhang, Lance and Parikh, Adeet and Lo, Aaron and Joshi, Abhishek and Mandlekar, Ajay and Zhu, Yuke},
  journal={arXiv preprint arXiv:2406.02523},
  year={2024}
}

@misc{mandlekar2023mimicgendatagenerationscalable,
      title={MimicGen: A Data Generation System for Scalable Robot Learning using Human Demonstrations}, 
      author={Ajay Mandlekar and Soroush Nasiriany and Bowen Wen and Iretiayo Akinola and Yashraj Narang and Linxi Fan and Yuke Zhu and Dieter Fox},
      year={2023},
      eprint={2310.17596},
      archivePrefix={arXiv},
      primaryClass={cs.RO},
      url={https://arxiv.org/abs/2310.17596}, 
}

@misc{liu2023liberobenchmarkingknowledgetransfer,
      title={LIBERO: Benchmarking Knowledge Transfer for Lifelong Robot Learning}, 
      author={Bo Liu and Yifeng Zhu and Chongkai Gao and Yihao Feng and Qiang Liu and Yuke Zhu and Peter Stone},
      year={2023},
      eprint={2306.03310},
      archivePrefix={arXiv},
      primaryClass={cs.AI},
      url={https://arxiv.org/abs/2306.03310}, 
}

@misc{mees2022calvinbenchmarklanguageconditionedpolicy,
      title={CALVIN: A Benchmark for Language-Conditioned Policy Learning for Long-Horizon Robot Manipulation Tasks}, 
      author={Oier Mees and Lukas Hermann and Erick Rosete-Beas and Wolfram Burgard},
      year={2022},
      eprint={2112.03227},
      archivePrefix={arXiv},
      primaryClass={cs.RO},
      url={https://arxiv.org/abs/2112.03227}, 
}

@inproceedings{yu2020meta,
  title={Meta-world: A benchmark and evaluation for multi-task and meta reinforcement learning},
  author={Yu, Tianhe and Quillen, Deirdre and He, Zhanpeng and Julian, Ryan and Hausman, Karol and Finn, Chelsea and Levine, Sergey},
  booktitle={Conference on robot learning},
  pages={1094--1100},
  year={2020},
  organization={PMLR}
}

@misc{mu2021maniskillgeneralizablemanipulationskill,
      title={ManiSkill: Generalizable Manipulation Skill Benchmark with Large-Scale Demonstrations}, 
      author={Tongzhou Mu and Zhan Ling and Fanbo Xiang and Derek Yang and Xuanlin Li and Stone Tao and Zhiao Huang and Zhiwei Jia and Hao Su},
      year={2021},
      eprint={2107.14483},
      archivePrefix={arXiv},
      primaryClass={cs.LG},
      url={https://arxiv.org/abs/2107.14483}, 
}

@misc{chen2021learninggeneralizableroboticreward,
      title={Learning Generalizable Robotic Reward Functions from "In-The-Wild" Human Videos}, 
      author={Annie S. Chen and Suraj Nair and Chelsea Finn},
      year={2021},
      eprint={2103.16817},
      archivePrefix={arXiv},
      primaryClass={cs.RO},
      url={https://arxiv.org/abs/2103.16817}, 
}

@misc{zakka2021xirlcrossembodimentinversereinforcement,
      title={XIRL: Cross-embodiment Inverse Reinforcement Learning}, 
      author={Kevin Zakka and Andy Zeng and Pete Florence and Jonathan Tompson and Jeannette Bohg and Debidatta Dwibedi},
      year={2021},
      eprint={2106.03911},
      archivePrefix={arXiv},
      primaryClass={cs.RO},
      url={https://arxiv.org/abs/2106.03911}, 
}

@misc{smith2020avidlearningmultistagetasks,
      title={AVID: Learning Multi-Stage Tasks via Pixel-Level Translation of Human Videos}, 
      author={Laura Smith and Nikita Dhawan and Marvin Zhang and Pieter Abbeel and Sergey Levine},
      year={2020},
      eprint={1912.04443},
      archivePrefix={arXiv},
      primaryClass={cs.RO},
      url={https://arxiv.org/abs/1912.04443}, 
}

@misc{sermanet2018timecontrastivenetworksselfsupervisedlearning,
      title={Time-Contrastive Networks: Self-Supervised Learning from Video}, 
      author={Pierre Sermanet and Corey Lynch and Yevgen Chebotar and Jasmine Hsu and Eric Jang and Stefan Schaal and Sergey Levine},
      year={2018},
      eprint={1704.06888},
      archivePrefix={arXiv},
      primaryClass={cs.CV},
      url={https://arxiv.org/abs/1704.06888}, 
}

@misc{eze2025learningwatchingreviewvideobased,
      title={Learning by Watching: A Review of Video-based Learning Approaches for Robot Manipulation}, 
      author={Chrisantus Eze and Christopher Crick},
      year={2025},
      eprint={2402.07127},
      archivePrefix={arXiv},
      primaryClass={cs.RO},
      url={https://arxiv.org/abs/2402.07127}, 
}

@article{feng2026human,
  title={From Human Videos to Robot Manipulation: A Survey on Scalable Vision-Language-Action Learning with Human-Centric Data},
  author={Feng, Zhiyuan and Li, Qixiu and Liang, Huizhi and Yang, Rushuai and Shen, Yichao and Du, Zhiying and Zhang, Zhaowei and Deng, Yu and Zhao, Li and Zhao, Hao and others},
  year={2026},
  publisher={TechRxiv}
}

@inproceedings{zhang2025rewind,
  title={ReWiND: Learning New Tasks from Language Without New Demonstrations},
  author={Zhang, Jiahui and Luo, Yusen and Anwar, Abrar and Sontakke, Sumedh Anand and Lim, Joseph J and Thomason, Jesse and Biyik, Erdem and Zhang, Jesse},
  booktitle={Inductive Biases in Reinforcement Learning Workshop@ RLC 2025},
  year={2025}
}

@inproceedings{yang2024adapt2reward,
  title={Adapt2reward: Adapting video-language models to generalizable robotic rewards via failure prompts},
  author={Yang, Yanting and Chen, Minghao and Qiu, Qibo and Wu, Jiahao and Wang, Wenxiao and Lin, Binbin and Guan, Ziyu and He, Xiaofei},
  booktitle={European Conference on Computer Vision},
  pages={163--180},
  year={2024},
  organization={Springer}
}

@inproceedings{ayalew2025progressor,
  title={Progressor: A perceptually guided reward estimator with self-supervised online refinement},
  author={Ayalew, Tewodros W and Zhang, Xiao and Wu, Kevin Yuanbo and Jiang, Tianchong and Maire, Michael and Walter, Matthew R},
  booktitle={Proceedings of the IEEE/CVF International Conference on Computer Vision},
  pages={10297--10306},
  year={2025}
}

@article{bu2025univla,
  title={Univla: Learning to act anywhere with task-centric latent actions},
  author={Bu, Qingwen and Yang, Yanting and Cai, Jisong and Gao, Shenyuan and Ren, Guanghui and Yao, Maoqing and Luo, Ping and Li, Hongyang},
  journal={arXiv preprint arXiv:2505.06111},
  year={2025}
}

@article{ye2024latent,
  title={Latent action pretraining from videos},
  author={Ye, Seonghyeon and Jang, Joel and Jeon, Byeongguk and Joo, Sejune and Yang, Jianwei and Peng, Baolin and Mandlekar, Ajay and Tan, Reuben and Chao, Yu-Wei and Lin, Bill Yuchen and others},
  journal={arXiv preprint arXiv:2410.11758},
  year={2024}
}

@article{majumdar2023we,
  title={Where are we in the search for an artificial visual cortex for embodied intelligence?},
  author={Majumdar, Arjun and Yadav, Karmesh and Arnaud, Sergio and Ma, Jason and Chen, Claire and Silwal, Sneha and Jain, Aryan and Berges, Vincent-Pierre and Wu, Tingfan and Vakil, Jay and others},
  journal={Advances in Neural Information Processing Systems},
  volume={36},
  pages={655--677},
  year={2023}
}

@article{ma2022vip,
  title={Vip: Towards universal visual reward and representation via value-implicit pre-training},
  author={Ma, Yecheng Jason and Sodhani, Shagun and Jayaraman, Dinesh and Bastani, Osbert and Kumar, Vikash and Zhang, Amy},
  journal={arXiv preprint arXiv:2210.00030},
  year={2022}
}

@article{nair2022r3m,
  title={R3m: A universal visual representation for robot manipulation},
  author={Nair, Suraj and Rajeswaran, Aravind and Kumar, Vikash and Finn, Chelsea and Gupta, Abhinav},
  journal={arXiv preprint arXiv:2203.12601},
  year={2022}
}

@article{black2024pi_0,
  title={$\pi_{0} $: A Vision-Language-Action Flow Model for General Robot Control},
  author={Black, Kevin and Brown, Noah and Driess, Danny and Esmail, Adnan and Equi, Michael and Finn, Chelsea and Fusai, Niccolo and Groom, Lachy and Hausman, Karol and Ichter, Brian and others},
  journal={arXiv preprint arXiv:2410.24164},
  year={2024}
}

@article{kim2024openvla,
  title={Openvla: An open-source vision-language-action model},
  author={Kim, Moo Jin and Pertsch, Karl and Karamcheti, Siddharth and Xiao, Ted and Balakrishna, Ashwin and Nair, Suraj and Rafailov, Rafael and Foster, Ethan and Lam, Grace and Sanketi, Pannag and others},
  journal={arXiv preprint arXiv:2406.09246},
  year={2024}
}

@inproceedings{bharadhwaj2024roboagent,
  title={Roboagent: Generalization and efficiency in robot manipulation via semantic augmentations and action chunking},
  author={Bharadhwaj, Homanga and Vakil, Jay and Sharma, Mohit and Gupta, Abhinav and Tulsiani, Shubham and Kumar, Vikash},
  booktitle={2024 IEEE International Conference on Robotics and Automation (ICRA)},
  pages={4788--4795},
  year={2024},
  organization={IEEE}
}

@inproceedings{zhang2025act,
  title={ACT-TSA: Action Chunking Transformer with Two-Stage Attention for Temporal Multimodality in Bimanual Manipulation Tasks},
  author={Zhang, Jiafan},
  booktitle={2025 IEEE 7th International Conference on Power, Intelligent Computing and Systems (ICPICS)},
  pages={764--772},
  year={2025},
  organization={IEEE}
}

@article{george2023one,
  title={One act play: Single demonstration behavior cloning with action chunking transformers},
  author={George, Abraham and Farimani, Amir Barati},
  journal={arXiv preprint arXiv:2309.10175},
  year={2023}
}

@article{fu2024mobile,
  title={Mobile aloha: Learning bimanual mobile manipulation with low-cost whole-body teleoperation},
  author={Fu, Zipeng and Zhao, Tony Z and Finn, Chelsea},
  journal={arXiv preprint arXiv:2401.02117},
  year={2024}
}

@article{zhao2024aloha,
  title={Aloha unleashed: A simple recipe for robot dexterity},
  author={Zhao, Tony Z and Tompson, Jonathan and Driess, Danny and Florence, Pete and Ghasemipour, Kamyar and Finn, Chelsea and Wahid, Ayzaan},
  journal={arXiv preprint arXiv:2410.13126},
  year={2024}
}

@article{chi2025diffusion,
  title={Diffusion policy: Visuomotor policy learning via action diffusion},
  author={Chi, Cheng and Xu, Zhenjia and Feng, Siyuan and Cousineau, Eric and Du, Yilun and Burchfiel, Benjamin and Tedrake, Russ and Song, Shuran},
  journal={The International Journal of Robotics Research},
  volume={44},
  number={10-11},
  pages={1684--1704},
  year={2025},
  publisher={Sage Publications Sage UK: London, England}
}

@article{shafiullah2023bringing,
  title={On bringing robots home},
  author={Shafiullah, Nur Muhammad Mahi and Rai, Anant and Etukuru, Haritheja and Liu, Yiqian and Misra, Ishan and Chintala, Soumith and Pinto, Lerrel},
  journal={arXiv preprint arXiv:2311.16098},
  year={2023}
}

@article{bousmalis2023robocat,
  title={Robocat: A self-improving generalist agent for robotic manipulation},
  author={Bousmalis, Konstantinos and Vezzani, Giulia and Rao, Dushyant and Devin, Coline and Lee, Alex X and Bauz{\'a}, Maria and Davchev, Todor and Zhou, Yuxiang and Gupta, Agrim and Raju, Akhil and others},
  journal={arXiv preprint arXiv:2306.11706},
  year={2023}
}

@article{team2024octo,
  title={Octo: An open-source generalist robot policy},
  author={Team, Octo Model and Ghosh, Dibya and Walke, Homer and Pertsch, Karl and Black, Kevin and Mees, Oier and Dasari, Sudeep and Hejna, Joey and Kreiman, Tobias and Xu, Charles and others},
  journal={arXiv preprint arXiv:2405.12213},
  year={2024}
}

@inproceedings{o2024open,
  title={Open x-embodiment: Robotic learning datasets and rt-x models: Open x-embodiment collaboration 0},
  author={O’Neill, Abby and Rehman, Abdul and Maddukuri, Abhiram and Gupta, Abhishek and Padalkar, Abhishek and Lee, Abraham and Pooley, Acorn and Gupta, Agrim and Mandlekar, Ajay and Jain, Ajinkya and others},
  booktitle={2024 IEEE International Conference on Robotics and Automation (ICRA)},
  pages={6892--6903},
  year={2024},
  organization={IEEE}
}

@article{jiang2023vima,
  title={Vima: Robot manipulation with multimodal prompts},
  author={Jiang, Yunfan and Gupta, Agrim and Zhang, Zichen and Wang, Guanzhi and Dou, Yongqiang and Chen, Yanjun and Fei-Fei, Li and Anandkumar, Anima and Zhu, Yuke and Fan, Linxi},
  year={2023}
}

@article{driess2023palm,
  title={Palm-e: An embodied multimodal language model},
  author={Driess, Danny and Xia, Fei and Sajjadi, Mehdi SM and Lynch, Corey and Chowdhery, Aakanksha and Ichter, Brian and Wahid, Ayzaan and Tompson, Jonathan and Vuong, Quan and Yu, Tianhe and others},
  journal={arXiv preprint arXiv:2303.03378},
  year={2023}
}

@article{ahn2022can,
  title={Do as i can, not as i say: Grounding language in robotic affordances},
  author={Ahn, Michael and Brohan, Anthony and Brown, Noah and Chebotar, Yevgen and Cortes, Omar and David, Byron and Finn, Chelsea and Fu, Chuyuan and Gopalakrishnan, Keerthana and Hausman, Karol and others},
  journal={arXiv preprint arXiv:2204.01691},
  year={2022}
}

@inproceedings{zitkovich2023rt,
  title={Rt-2: Vision-language-action models transfer web knowledge to robotic control},
  author={Zitkovich, Brianna and Yu, Tianhe and Xu, Sichun and Xu, Peng and Xiao, Ted and Xia, Fei and Wu, Jialin and Wohlhart, Paul and Welker, Stefan and Wahid, Ayzaan and others},
  booktitle={Conference on Robot Learning},
  pages={2165--2183},
  year={2023},
  organization={PMLR}
}

@article{brohan2022rt,
  title={Rt-1: Robotics transformer for real-world control at scale},
  author={Brohan, Anthony and Brown, Noah and Carbajal, Justice and Chebotar, Yevgen and Dabis, Joseph and Finn, Chelsea and Gopalakrishnan, Keerthana and Hausman, Karol and Herzog, Alex and Hsu, Jasmine and others},
  journal={arXiv preprint arXiv:2212.06817},
  year={2022}
}

@article{BCZ,
  author       = {Eric Jang and
                  Alex Irpan and
                  Mohi Khansari and
                  Daniel Kappler and
                  Frederik Ebert and
                  Corey Lynch and
                  Sergey Levine and
                  Chelsea Finn},
  title        = {{BC-Z:} Zero-Shot Task Generalization with Robotic Imitation Learning},
  journal      = {CoRR},
  volume       = {abs/2202.02005},
  year         = {2022},
  url          = {https://arxiv.org/abs/2202.02005},
  eprinttype   = {arXiv},
  eprint       = {2202.02005},
  bibsource    = {dblp computer science bibliography, https://dblp.org}
}

@inproceedings{walke2023bridgedata,
  title={Bridgedata v2: A dataset for robot learning at scale},
  author={Walke, Homer Rich and Black, Kevin and Zhao, Tony Z and Vuong, Quan and Zheng, Chongyi and Hansen-Estruch, Philippe and He, Andre Wang and Myers, Vivek and Kim, Moo Jin and Du, Max and others},
  booktitle={Conference on Robot Learning},
  pages={1723--1736},
  year={2023},
  organization={PMLR}
}

@article{RoboNet,
  author       = {Sudeep Dasari and
                  Frederik Ebert and
                  Stephen Tian and
                  Suraj Nair and
                  Bernadette Bucher and
                  Karl Schmeckpeper and
                  Siddharth Singh and
                  Sergey Levine and
                  Chelsea Finn},
  title        = {RoboNet: Large-Scale Multi-Robot Learning},
  journal      = {CoRR},
  volume       = {abs/1910.11215},
  year         = {2019},
  url          = {http://arxiv.org/abs/1910.11215},
  eprinttype   = {arXiv},
  eprint       = {1910.11215},
  bibsource    = {dblp computer science bibliography, https://dblp.org}
}

\end{document}